\documentclass{article}
\ifdefined\pdfoutput\pdfoutput=1\fi

\usepackage[final]{galaxea}

\usepackage[utf8]{inputenc} 
\usepackage[T1]{fontenc}    
\usepackage{url}            
\usepackage{booktabs}       
\usepackage{amsfonts}       
\usepackage{graphicx}       
\usepackage{nicefrac}       
\usepackage{microtype}      
\usepackage{xcolor}         
\usepackage{amsmath, amssymb, bm}
\usepackage{algorithm}
\usepackage[noend]{algpseudocode}
\usepackage{algorithmicx}
\usepackage{array}
\usepackage{colortbl}
\usepackage{natbib}
\usepackage{caption}
\usepackage{hyperref}       
\usepackage{pifont}%
\usepackage{xspace}
\usepackage{enumitem}       
\usepackage{etoolbox}

\usepackage{graphics} %
\usepackage{epsfig} %
\usepackage{newpxtext} 
\usepackage{newpxmath} 
\usepackage{multirow}
\usepackage{makecell}

\usepackage{framed}
\usepackage{wrapfig}
\usepackage{titletoc}
\usepackage{subcaption, overpic, textpos}
\usepackage{titlesec}
\usepackage{comment}
\usepackage{fontawesome}
\usepackage{siunitx}
\usepackage{tcolorbox} %
\usepackage{cleveref}
\tcbuselibrary{listings} %

\paperlogo{\includegraphics[height=0.6cm]{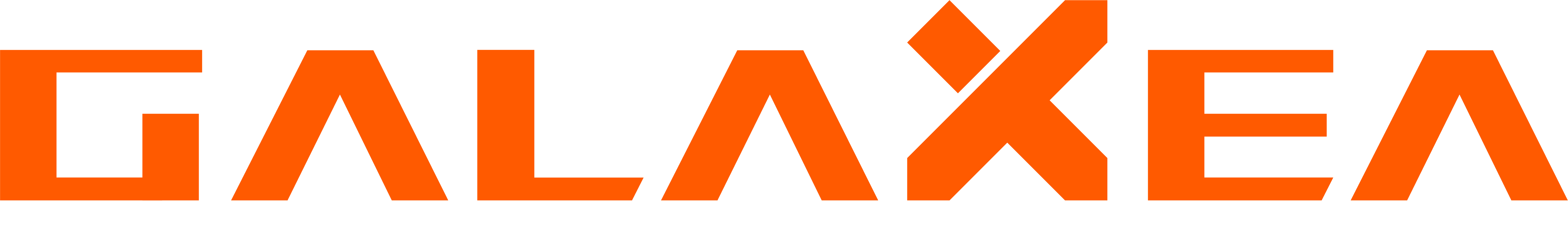}}

\definecolor{galaxeaOrange}{HTML}{E8590C}   
\definecolor{galaxeaDeep}{HTML}{C7480C}     
\definecolor{galaxeaBurnt}{HTML}{B3591F}    
\definecolor{refSienna}{HTML}{B3591F}       
\definecolor{citeTeal}{HTML}{3C4D57}        
\definecolor{urlSlate}{HTML}{46708A}        
\definecolor{codebg}{HTML}{F8F5EF}          
\definecolor{codeframe}{HTML}{DDD6CA}       
\newcommand{\terminalfamily}{\fontfamily{zi4}\selectfont} 
\hypersetup{colorlinks=true, citecolor=citeTeal, linkcolor=refSienna, urlcolor=urlSlate}
\newcommand{\best}[1]{\textcolor{galaxeaDeep}{\textbf{#1}}}

\ifdefined\XeTeXversion
  \usepackage{fontspec}
  \newfontfamily\GmManrope{Manrope-SemiBold.ttf}[Path=./fonts/manrope/]
  \DeclareRobustCommand{\Gm}{\mbox{{\GmManrope\color{galaxeaOrange}G0.5}}\xspace}
\else
  \DeclareRobustCommand{\Gm}{\mbox{\sffamily\bfseries\color{galaxeaOrange}G0.5}\xspace}
\fi
\newcommand{\RoneLite}{R1-Lite\xspace}
\newcommand{\RonePro}{R1-Pro\xspace}

\titleformat{\section}
  {\sffamily\bfseries\Large\color{galaxeaOrange}}  
  {\thesection}{1em}{}  

\titleformat{\subsection}
  {\sffamily\bfseries\large\color{galaxeaOrange}}
  {\thesubsection}{1em}{}

\titleformat{\subsubsection}
  {\sffamily\bfseries\normalsize\color{galaxeaOrange}}
  {\thesubsubsection}{1em}{}

\title{Galaxea G0.5: One Autoregressive Stream for Robot Reasoning and Action}

\author{%
  Galaxea Team \\
  \url{https://opengalaxea.github.io/G05/} \\
}

\usepackage{ifxetex}
\ifxetex
  \usepackage{xeCJK}
\fi

\begin{document}

\maketitle

\vspace{-0.6em}
\noindent\makebox[\textwidth][c]{\includegraphics[width=1.1\textwidth]{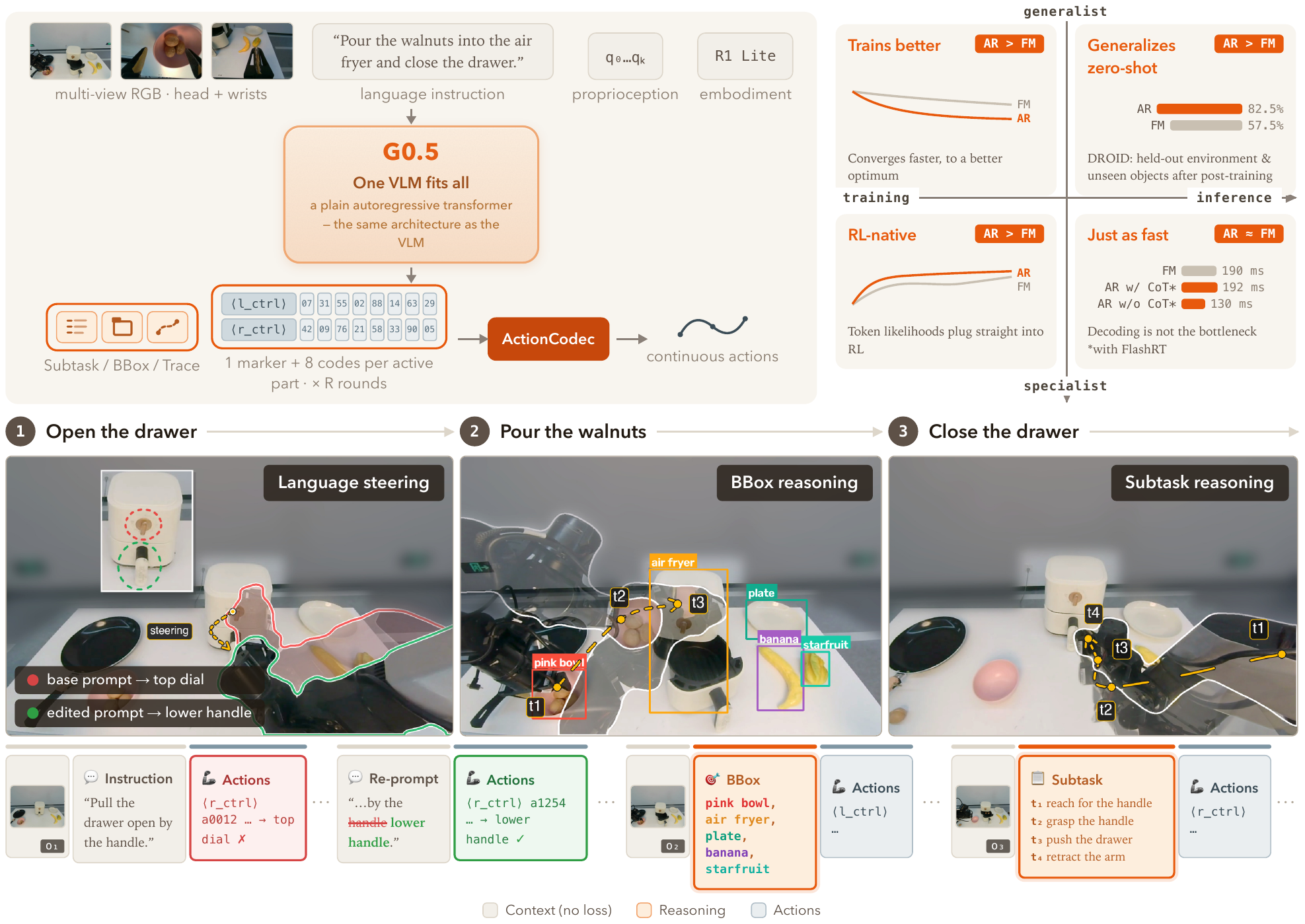}}
\begingroup
\captionsetup{font=small}
\captionof{figure}{\textbf{\Gm: reasoning and action in one autoregressive stream.}
\textbf{Top left:} A single VLM conditions on multi-view images, an instruction, proprioception, and an embodiment identifier, then generates optional chain-of-thought (CoT) and compact action codes under the same next-token objective. A cross-embodiment ActionCodec converts the codes into continuous motor commands, while active-part tokenization omits inactive control groups.
\textbf{Top right:} Comparisons with flow-matching policies span generalist pretraining and specialist post-training, covering optimization, compatibility with reinforcement learning, transfer after DROID post-training to a held-out physical setup and previously unseen object instances, and inference latency.
\textbf{Bottom:} A closed-loop rollout of \emph{open the drawer $\rightarrow$ pour the walnuts $\rightarrow$ close the drawer}. At each stage, updated visual and proprioceptive observations condition interleaved reasoning and action generation. The examples highlight prompt-driven action steering (left), grounded bounding-box prediction (middle), and adaptive subtask reasoning (right); colors distinguish conditioning context, reasoning, and action tokens.}
\label{fig:teaser}
\endgroup
\newpage

\begin{abstract}
The prevailing recipe for Vision-Language-Action (VLA) models couples a pretrained VLM with a separately trained flow-matching action expert. This makes the VLM a context encoder rather than a decision-maker.
Instead, we argue for focusing on the VLM backbone: a unified model with a single set of weights that generates both reasoning and actions within a single autoregressive token stream.

We introduce \Gm, a pretrained autoregressive VLA in which a single transformer decoder emits reasoning and action tokens under a single objective. Three components make this tractable at foundation-model scale: a learnable cross-embodiment action tokenizer that maps heterogeneous robot actions into a shared vocabulary; a native chain-of-thought stream interleaving task decomposition, object grounding, and action hints with action tokens; and a visual memory module that injects multi-second history through the vision encoder. Because reasoning and action share a single set of weights, the pretrained VLM's capabilities carry over to physical behavior: the model follows instructions closely, and prompts directly steer action granularity, task horizon, and out-of-distribution scene handling without further training.
Pretrained on a large collection of robot datasets together with VQA samples, \Gm surpasses state-of-the-art models across 7 independent regimes: real-world fine-tuning on \RoneLite/\RonePro robots (76.7\% vs.\ 53.3\% for $\pi_{0.5}$ and 24.4\% for GR00T-N1.7), the 2025 BEHAVIOR Challenge on 50 long-horizon household mobile manipulation tasks using a generalist policy (31.4\% vs.\ 26.3\% for $\pi_{0.5}$ and 26.1\% for the challenge winner), DROID post-training followed by zero-shot transfer to an unseen environment and objects (82.5\%), a language-following Pick-and-Place benchmark, LIBERO (98.9\%), RoboTwin 2.0 (93.3\%), and SimplerEnv-Bridge (87.3\%).
\end{abstract}

\section{Introduction}

Vision-Language-Action (VLA) models have rapidly emerged as a leading paradigm for general-purpose robot control, extending large-scale vision-language pretraining from perception and language understanding to physical action \citep{rt2, openvla, pi0}. Early VLA systems used an autoregressive interface that cast robot control as token generation: continuous actions were discretized, appended to the language vocabulary, and predicted by the VLM alongside text tokens \citep{rt2, openvla}. This keeps the VLM itself as the actor, but scales poorly. As control frequency, action horizon, and action dimensionality increase, per-timestep autoregressive action tokens grow rapidly, making high-frequency control slow and expensive. This bottleneck pushed the field toward VLM-as-encoder architectures, where a pretrained VLM supplies hidden states or KV cache to a separately trained flow-matching or diffusion expert that predicts continuous action chunks \citep{pi0, pi0.5, gr00t-n1, smolvla}.

This shift improves action efficiency, but changes the role of the VLM. In VLM-as-encoder models, the VLM is no longer the action generator; it becomes a vision-language condition encoder, while the final action distribution is produced by an expert with separate parameters and a separate objective. Consequently, core generative capabilities of VLMs---chain-of-thought reasoning, in-context learning, and prompt-based motion steering---can affect behavior only after passing through a compressed conditioning bottleneck, rather than as native parts of action generation \citep{cot-vla, dualcot-vla, halo}. We therefore return to the autoregressive formulation, but remove the source of its original inefficiency: excessive action tokenization. A learning-based VQ tokenizer compresses action chunks into compact discrete codes, while active degree-of-freedom prediction avoids spending tokens on robot joints that do not need to move. Together, these choices substantially reduce the decoding burden while preserving the VLM as a generative actor. As part of the pretrained backbone, we also retain a lightweight visual-memory mechanism that feeds accumulated visual context through the vision encoder, following recent memory-augmented VLA designs \citep{pi-mem}, since persistent visual context benefits long-horizon control and closed-loop replanning. More importantly, once reasoning and action share the same autoregressive stream, chain-of-thought can be trained as a native component of control: the model can zero-shot decompose an instruction into subtasks, identify task-relevant objects and their bounding boxes, and feed these intermediate predictions directly into subsequent action generation (Fig.~\ref{fig:teaser}).

We introduce \Gm, a pretrained autoregressive VLA in which a single model reasons, plans, and acts within a unified token stream spanning images, language, reasoning traces, and actions. Our contributions are as follows:
\begin{figure}[!tbp]
\centering
\includegraphics[width=\linewidth]{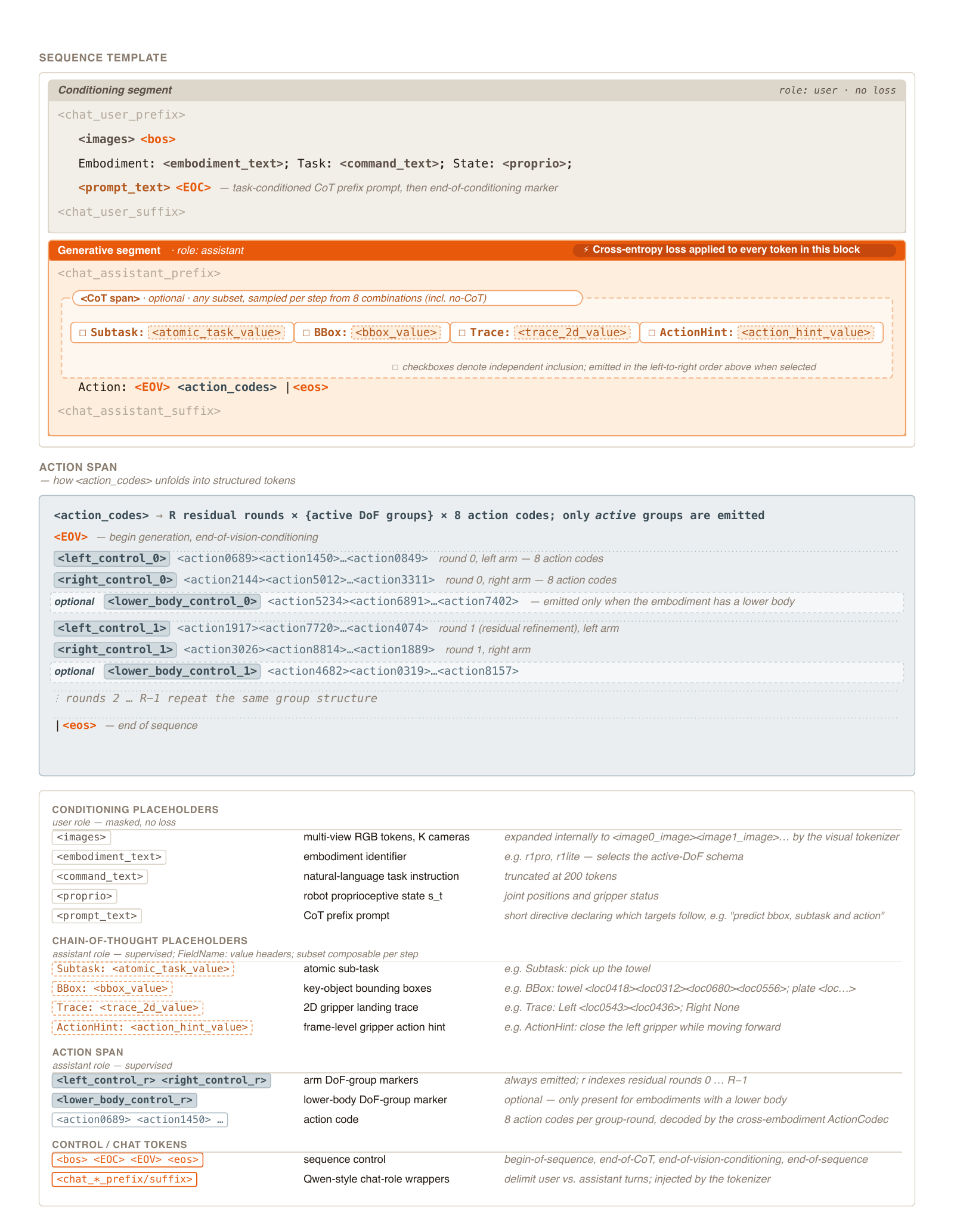}
\vspace{-2em}
\caption{\textbf{Token sequence template.} All inputs and outputs are serialised into a single autoregressive sequence: a \textit{conditioning segment} (multi-view RGB, embodiment id, task instruction, proprioceptive state---in user-side chat tokens) and a \textit{generative segment} on which the next-token cross-entropy loss in Eq.~\eqref{eq:loss} is applied. The generative segment composes an optional chain-of-thought span---any subset of four self-describing reasoning targets (\texttt{Subtask:}, \texttt{BBox:}, \texttt{Trace:}, \texttt{ActionHint:})---followed by the action codes, which themselves expand into $R$ residual rounds of DoF-group markers each followed by 8 action codes (Sec.~\ref{sec:codec}).}
\label{fig:token-template}
\end{figure}

\textbf{(1) A unified heterogeneous action codec.} We pretrain a learning-based action codec that maps continuous action sequences from embodiments with different degrees of freedom, control frequencies, and morphologies into a shared token vocabulary. Unlike FAST, which applies a fixed DCT-based pipeline separately to each embodiment \citep{fast}, our codec is learned end-to-end and cross-embodiment by design. It allows the VLM to represent actions from different robots through a common discrete interface, making autoregressive VLA practical at foundation-model scale; Fig.~\ref{fig:teaser} illustrates this on \RoneLite, where the active-token layout adapts on the fly to whichever parts are in motion and drops the idle arm's token group from the stream entirely rather than padding it.
\textbf{(2) Native chain-of-thought through autoregressive training.} We construct a family of CoT templates for task decomposition, scene grounding, and sub-goal sequencing, and train the model to emit reasoning tokens before and between action tokens in the same autoregressive stream. Unlike CoT-VLA, DualCoT-VLA, and related approaches that attach reasoning modules to VLM-as-encoder backbones \citep{cot-vla, dualcot-vla, flowvla, halo}, our CoT tokens share the decoder, context, and objective with the action tokens. Reasoning and action are therefore not separate stages, but coupled phases of one generative process (see the interleaved CoT and action segments in Fig.~\ref{fig:teaser}). This design yields two benefits that we evaluate separately: stronger grounding and execution under long-horizon instructions---including zero-shot execution of household tasks under stage-conditioned instructions outside the pretraining distribution (Sec.~\ref{subsec:cot_head})---and improved language following beyond what the codec alone provides.
\textbf{(3) Emergent prompt-driven behavior control.} Preserving the autoregressive interface keeps the VLM's in-context language capacity directly wired to action generation, in principle enabling prompt-level steering of physical behavior without retraining. In our zero-shot probes (Sec.~\ref{subsec:cot_head}) we see preliminary qualitative indications of this---per-stage instruction wording such as adverbial qualifiers, spatial cues, or near-synonymous verbs visibly shifts policy behavior---and leave a systematic study to future work. We suspect this capacity is partly structural to the autoregressive interface: when the VLM only conditions an external expert, prompts can shape the condition but cannot directly reshape the next-action distribution.

We evaluate \Gm across seven settings that probe distinct facets of a general-purpose VLA: real-world fine-tuning on the \RoneLite and \RonePro bimanual platforms, the BEHAVIOR-1K Challenge on 50 long-horizon household mobile-manipulation tasks~\citep{li2023behavior}, DROID post-training with zero-shot transfer to an unseen environment and objects, a Pick-and-Place language-following benchmark, and three standardized simulation suites (LIBERO, RoboTwin~2.0, SimplerEnv-Bridge)~\citep{libero, simplerenv}. We compare against representative baselines from three model families: VLM-as-encoder models \citep{pi0, pi0.5, gr00t-n1, smolvla}, autoregressive models \citep{openvla, fast}, and the recently popular world action models \citep{fastwam}. Three findings stand out. First, on standard task success metrics, \Gm matches or surpasses the strongest baselines across these families---98.9\% on LIBERO, 93.3\% on RoboTwin~2.0, 87.3\% on SimplerEnv-Bridge, 82.5\% on DROID with environment- and object-level zero-shot transfer, and 76.7\% on the \RoneLite and \RonePro platforms, compared with 53.3\% for $\pi_{0.5}$ and 24.4\% for GR00T-N1.7---indicating that the pretrained \Gm backbone transfers effectively to downstream control across these heterogeneous suites. Second, on language following and multi-stage execution under stage-conditioned prompts, \Gm substantially outperforms VLM-as-encoder baselines on the Pick-and-Place benchmark and the BEHAVIOR-1K Challenge, where a single \Gm checkpoint trained for one post-training epoch already surpasses both $\pi_{0.5}$ trained for four epochs and the four-checkpoint Challenge winner. This is consistent with our argument that these capabilities are structurally weakened when the VLM is reduced to a condition encoder. Third, a small qualitative probe on two zero-shot long-horizon household tasks (Sec.~\ref{subsec:cot_head}) suggests that prompt wording---adverbial qualifiers, spatial cues, and verb substitutions---can shift AR$+$CoT rollouts without retraining; we report this as an early hook for prompt-level behavior steering rather than a quantitative claim, and defer a systematic study to future work. Taken together, our results suggest that the path forward for VLA is not to place increasingly sophisticated action experts on top of an underused VLM, but to let the VLM remain what pretraining made it: an autoregressive reasoner that can also act, remember, and adapt in context. We hope this work re-establishes autoregressive modeling as a foundation for VLA and that the pretrained backbone we release provides a useful starting point for future work.

\section{Related Work}
\label{sec:related}
\subsection{VLA Architectures: from VLM-as-Encoder to VLM-as-Actor}
\label{sec:related-arch}

Vision-language-action models split along one architectural axis: whether the VLM produces actions or only conditions a separate module that does. The dominant line couples a pretrained VLM with an action expert that consumes its features and emits continuous actions via diffusion or flow matching: $\pi_0$~\citep{pi0} introduced a separately-parameterized expert with block-wise causal attention, and $\pi_{0.5}$~\citep{pi0.5}, GR00T-N1 / N1.5 / N1.6~\citep{gr00t-n1}, and SmolVLA~\citep{smolvla} follow variants of the same template. The autoregressive line, including RT-2~\citep{rt2}, OpenVLA~\citep{openvla}, and $\pi_0$-FAST~\citep{fast}, instead discretizes actions and predicts them with the VLM itself under next-token prediction. The two lines are typically presented as a trade-off---continuous heads for smooth high-frequency control, AR for reasoning and simplicity---but they also differ in what the VLM is \emph{for}: in the first line, the VLM is a condition encoder whose pretrained reasoning is exercised only indirectly, while in the second, it remains the agent that acts.

A revealing thread within the VLM-as-encoder line is the anti-forgetting problem: when the action expert's gradients flow back into the VLM, the VLM's pretrained perception and language capabilities degrade~\citep{pi05-KI, vla-0}. The mainstream remedy, Knowledge Insulation~\citep{pi05-KI}, stops these gradients and reintroduces AR action prediction as an auxiliary representation-learning objective for the backbone---implicitly conceding that AR action supervision is exactly the signal that protects the VLM's capabilities. Recent results push further: VLA-0~\citep{vla-0} shows that an unmodified VLM trained AR on actions-as-text outperforms $\pi_{0.5}$-KI, OpenVLA-OFT, and SmolVLA on LIBERO without large-scale action pretraining, providing direct evidence that the AR paradigm is not the bottleneck. Our work takes this signal seriously and commits to the AR line end-to-end, retaining a flow-matching head only as an optional inference accelerator. What remains open after VLA-0---and what the rest of this section traces---is how to scale the AR paradigm beyond a single low-frequency embodiment with closed-vocabulary tasks: through a tokenizer that respects morphological structure (Sec.~\ref{sec:related-tokenizer}) and reasoning that grounds language in action (Sec.~\ref{sec:related-cot}).

\subsection{Action Tokenization and Cross-Embodiment}
\label{sec:related-tokenizer}

Action tokenization for VLAs has progressed through three generations. Per-dimension, per-timestep binning, as in RT-2~\citep{rt2} and OpenVLA~\citep{openvla}, fails on high-frequency dexterous data because adjacent timesteps are strongly correlated and binning wastes capacity~\citep{fast}. FAST and FAST+~\citep{fast} replace binning with DCT plus byte-pair encoding, exploiting that correlation as compressible signal, and FAST+ is trained on one million trajectories to serve as a universal tokenizer. Neural and vector-quantized variants---VQ-VLA~\citep{vq-vla}, BEAST~\citep{beast}, and earlier VQ-BeT~\citep{vq-bet}---push reconstruction quality further at the cost of joint training and more complex pipelines.

Cross-embodiment generalization is largely orthogonal to all three. Mainstream VLAs handle morphological heterogeneity at the action-space level rather than the tokenizer: $\pi_0$~\citep{pi0} pads all robots to an 18-dim union state, GR00T-N1~\citep{gr00t-n1} uses per-embodiment MLP encoders and decoders, and SpatialVLA~\citep{spatialvla} unifies action spaces via adaptive grids. The closest neighbors to our work are Being-H0.5~\citep{being-h05}, which maps heterogeneous robot controls into semantically aligned slots and even folds the MANO hand model into the same scheme, Green-VLA~\citep{green-vla}, which retargets across robots by aligning corresponding parts into a unified action space, and HEX~\citep{hex}, whose humanoid-aligned state representation operates on canonical body-part abstractions. All three operate at the action-vector level. Our contribution is to lift the same structural alignment into the \emph{tokenizer itself}: a single frozen codec consumes a 5-part fixed-dimensional layout and emits a unified 27-dim action token stream, so left/right symmetry is preserved by construction and adding a new embodiment requires no new parameters in either the tokenizer or the action head.

\subsection{Reasoning and Chain-of-Thought in VLAs}
\label{sec:related-cot}

Two families have emerged for injecting reasoning into VLAs. \emph{Bolt-on} CoT routes natural-language plans or 2D paths from a high-level VLM into a separate low-level controller, as in HAMSTER~\citep{hamster} and Fast-in-Slow style System-2-feeds-System-1 designs~\citep{gr00t-n1}; the reasoning is an interface between modules rather than a co-generated component of the action. \emph{In-stream} CoT, by contrast, generates reasoning and action in the same AR sequence from the same decoder. ECoT~\citep{ecot} reports a 28-point absolute improvement on OpenVLA by training it to predict plans, subtasks, motions, bounding boxes, and end-effector positions before actions; CoT-VLA~\citep{cot-vla} replaces text reasoning with autoregressively-generated subgoal images; Emma-X~\citep{emma-x} predicts look-ahead 2D gripper checkpoints; and $\pi_{0.5}$~\citep{pi0.5} emits high-level subtask text from the VLM before invoking its flow-matching expert. Our setting is closest in spirit to ECoT in that reasoning and action share a single AR decoder, but differs along two axes that matter for our claims: we combine four reasoning primitives---object bounding boxes, atomic subtask text, 2D end-effector traces inspired by TraceVLA~\citep{trace-vla}, and action hints---in one shared token vocabulary, and we expose them as \emph{prompt-conditional} templates, letting the CoT mode be switched at inference without retraining.



\section{G0.5 Model Design}
\label{sec:method}

We design our model around a single commitment: perception, reasoning, and action should be unified within a single autoregressive process over a shared token vocabulary. This commitment shapes every component below---the action representation, the reasoning scaffold, the visual conditioning, and the training objective---and distinguishes our design from VLM-as-encoder architectures in which action generation lives in a separate module with a separate objective.

Our model is initialized from Qwen3.5 2B~\citep{qwen3.5}, a pretrained vision-language model that provides a strong visual encoder, a shared multimodal token vocabulary, and an autoregressive decoder. At inference, given (i) a short temporal window of multi-view RGB observations $\{o_{t-h}^{(k)}\}$ from $K$ cameras, (ii) an embodiment identifier $e$ (e.g., \RonePro), (iii) a natural-language task instruction $\ell$, and (iv) a proprioceptive state $s_t$, the model autoregressively generates a structured output that concludes in a sequence of discrete action codes. Depending on the prompt template, the generation can optionally be preceded by chain-of-thought (CoT) segments that ground objects, decompose subtasks, or sketch gripper traces. The action codes are decoded by our cross-embodiment ActionCodec into continuous control commands in a unified action space shared across embodiments. The autoregressive VLM is self-contained and serves as the default policy in all main experiments. For comparison and optional deployment, we additionally attach a flow-matching head that follows the action-expert architecture of $\pi_{0.5}$~\citep{pi0.5} and conditions on the autoregressive trunk's hidden states.

All inputs and outputs are serialized into a single token sequence following the template in Fig.~\ref{fig:token-template}. The sequence is partitioned into a \emph{conditioning segment}---wrapping images, embodiment, task, and state in user-side chat tokens and terminated by \texttt{<EOC>}---and a \emph{generative segment}---wrapping the CoT trace and action codes in assistant-side chat tokens, with \texttt{<EOV>} marking the boundary between reasoning and action emission. Training uses the standard next-token cross-entropy loss, computed only over the generative segment:
\begin{equation}
\mathcal{L}(\theta) \;=\; -\!\!\sum_{i \in \mathcal{G}} \log p_\theta\bigl(x_i \mid x_{<i}\bigr),
\label{eq:loss}
\end{equation}
where $\mathcal{G}$ indexes the generative-segment tokens. Crucially, this single loss jointly supervises CoT generation and action generation: there is no auxiliary regression objective or expert distillation in pre-training. CoT traces and actions are all ``just tokens'' to the decoder, drawn from the same vocabulary and produced by the same forward pass.

The remainder of this section unpacks the three components of the generative segment in the order they were derived: the cross-embodiment action codec (Sec.~\ref{sec:codec}), the chain-of-thought scaffold (Sec.~\ref{sec:cot}), and short-term visual memory (Sec.~\ref{sec:visual_memory}).

\subsection{Structured Tokenization of Heterogeneous Action Data}
\label{sec:codec}
A key challenge is how to represent heterogeneous actions from diverse embodiments in a structured token space that VLMs can efficiently model. Existing approaches suffer from two major limitations: \textbf{(a) lack of structural decomposition.} Most methods flatten the entire action space into a single vector before discretization \citep{fast,belkhale2024minivla,vq-vla,dong2026actioncodec}, regardless of embodiment topology or controllable degrees of freedom (DoFs). This results in semantically entangled action tokens that transfer poorly across embodiments. In addition, token count scales directly with the total number of controllable DoFs, despite the fact that only a small subset of joints are typically active at each timestep. \textbf{(b) poor token consistency.} Discrete action spaces are usually learned without explicit structural constraints, causing semantically similar actions to map to token sequences with large Hamming distances \citep{dong2026actioncodec}. As supervision signals for VLM training, such inconsistency introduces substantial optimization noise and reduces training efficiency.

\begin{figure}
    \centering
    \includegraphics[width=1.0\linewidth]{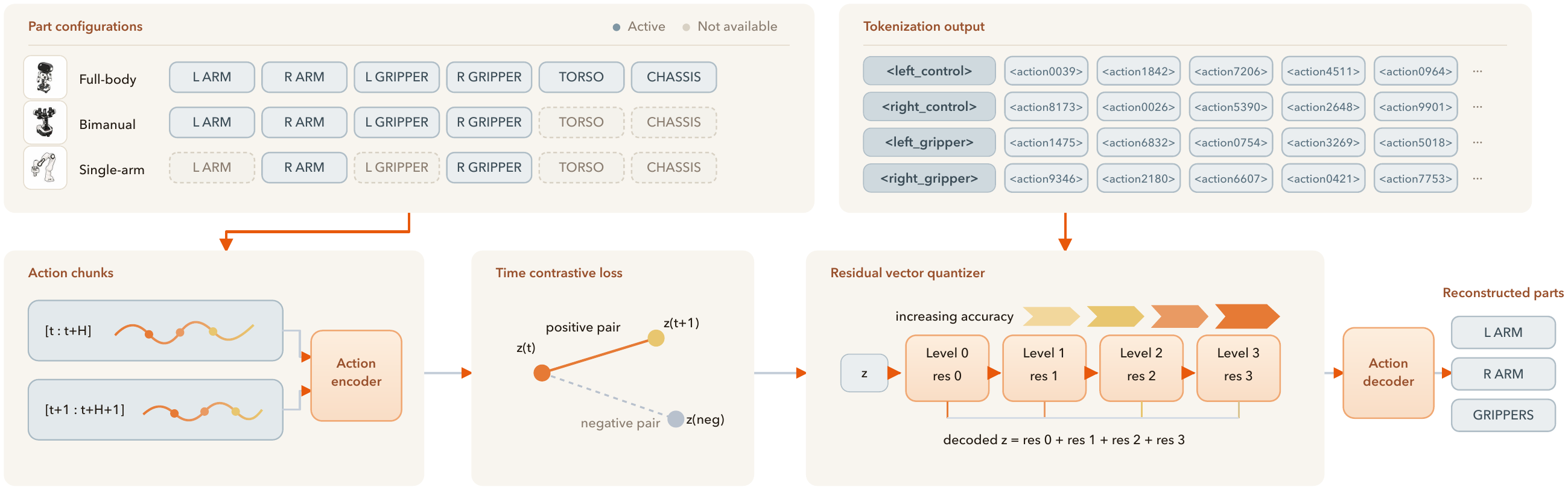}
    \caption{\textbf{Structured action tokenization.}
    Heterogeneous robot actions are decomposed into semantically aligned motion parts, encoded with a residual vector quantizer, and serialized as part-specific action tokens. This representation shares one action vocabulary across embodiments while allowing sparse prediction over only the activated parts.}
    \label{fig:action_tokenizer}
\end{figure}

To address these issues, we adopt the action grouping strategy of FASTer \citep{liu2026faster} together with the training recipe of ActionCodec \citep{dong2026actioncodec}. Specifically, we decompose each robot into independent motion parts (e.g., left control, right control, lower body), and pad each part to a shared maximum dimensionality before training a residual vector quantization (RVQ) model over the grouped actions. We further introduce a temporal contrastive objective to improve token consistency across temporally adjacent motions. During tokenization, we explicitly inject structural special tokens into the sequence. Concretely, the action span shown in the generative segment of Fig.~\ref{fig:token-template} unfolds into $R$ residual rounds, each containing the currently active DoF-group markers (\texttt{<left\_control\_r>}, \texttt{<right\_control\_r>}, and optionally \texttt{<lower\_body\_control\_r>} for embodiments with a lower body) followed by their 8 action codes. This formulation allows the model to predict only the motion parts that are actively involved in the current behavior. In practice, the proposed structured tokenization significantly improves training efficiency, enables heterogeneous embodiments to share a unified action configuration, and naturally supports sparse action prediction during inference, where inactive parts remain stationary without requiring additional token generation. We show the details in \Cref{fig:action_tokenizer}.

\subsection{Native Chain-of-Thought}
\label{sec:cot}
To preserve or further enhance the physical intelligence of VLMs, previous methods typically co-train auxiliary VQA tasks, such as sub-task or object bounding box prediction. However, these objectives are treated only as training-time supervision and never explicitly participate in the action generation process itself, making it difficult to directly assess whether such intermediate reasoning signals truly benefit downstream action prediction. In contrast, we leverage the unified autoregressive formulation of our model to naturally integrate these auxiliary tasks into the action generation stream as native chain-of-thought (CoT) reasoning. Instead of treating reasoning-related annotations as isolated supervision targets, the model is trained to optionally perform intermediate reasoning before action prediction across four self-describing targets---task decomposition (\texttt{Subtask:}), key-object localization (\texttt{BBox:}), motion planning (\texttt{Trace:}), and action hints (\texttt{ActionHint:})---which populate the CoT span shown in the generative segment of Fig.~\ref{fig:token-template}. Any subset of these targets can be emitted at each step, and we draw from 8 curated combinations (including a no-CoT baseline) per training step, all supervised within the same next-token objective.

Surprisingly, the resulting CoT capability exhibits strong zero-shot generalization. On unseen scenes and tasks, the model is able to generate accurate subtasks, proactively identify task-relevant objects together with their bounding boxes, and predict additional reasoning traces such as 2D motion trajectories and action hints. More importantly, enabling CoT reasoning consistently improves instruction-following behavior and action accuracy on complex manipulation tasks. These results suggest that intermediate reasoning is not merely an auxiliary supervision signal, but can serve as an effective test-time guidance for embodied action generation.

\subsection{Visual Memory}
\label{sec:visual_memory}
Complex mobile manipulation tasks are inherently non-Markovian. Relying solely on single frame observations often fails during temporary occlusions from robotic arms or environmental clutter, and lacks the temporal context needed to recognize failures and formulate alternative retry strategies. However, resolving partial observability by naively stacking historical vision tokens introduces severe limitations. It scales quadratically in computational cost, causing unacceptable latency for high-frequency control, and makes the model highly susceptible to error accumulation and state drifting when encountering unseen temporal trajectories.

To overcome these challenges, we follow $\pi_{0.7}$~\citep{pi07} and MEM~\citep{pi-mem} by inserting factorized spatial and temporal attention modules every four layers within the Vision Transformer. This separable mechanism efficiently fuses historical context by sequentially mixing information across time steps and spatial patches. To strictly bound computational latency, we discard all historical tokens at the final layer, and stochastically drop all historical frames during training to prevent overfitting. Finally, we replace discrete text tokenizers with continuous state embeddings to perfectly synchronize proprioceptive inputs with the corresponding visual frames.

\section{G0.5 Pre-training}
\label{sec:pretrain}

We pre-train \Gm in a single stage on a heterogeneous mixture of robot
demonstrations and web-scale vision--language data. The model, tokenization,
chain-of-thought (CoT) stream, and visual-memory module are described in Sec.~\ref{sec:method}; here we specify
only the data composition, the sampling and supervision recipe, and the
optimization setup.

\label{sec:pretrain-data}

\paragraph{Robot data mixture.}
The robot portion of the pre-training mixture covers \textbf{14 embodiments} across diverse real-world and simulated robot ontologies. DROID data are not part of this foundation pre-training mixture; for the evaluation in Sec.~\ref{subsec:droid_zeroshot}, the resulting model is subsequently post-trained on DROID data while excluding any demonstrations from the held-out evaluation environment and physical object instances.
All sources are cast into a single \textbf{27-dimensional unified action space},
partitioned as
\begin{tcolorbox}[colback=codebg, colframe=codeframe, boxrule=0.5pt, arc=3pt,
                  left=6pt, right=6pt, top=5pt, bottom=5pt,
                  before skip=8pt, after skip=8pt]
\centering
{\terminalfamily\small
<left\_control>(9)\,|\,<left\_gripper>(1)\,|\,<right\_control>(9)\,|\,<right\_gripper>(1)\,|\,<lower\_body>(7)}
\end{tcolorbox}
Slots that a given embodiment does not actuate are filled with \texttt{noop}
tokens at merge time, so embodiments of differing morphology share one output
head without per-robot adapters.
We treat each embodiment individually for pre-processing: the action
normalization mode (z-score with tail clipping, or $q_{01}/q_{99}$ quantile
scaling) and the per-channel action filters are set per source rather than
mixture-wide.

\begin{figure}[t]
\centering
\includegraphics[width=\columnwidth]{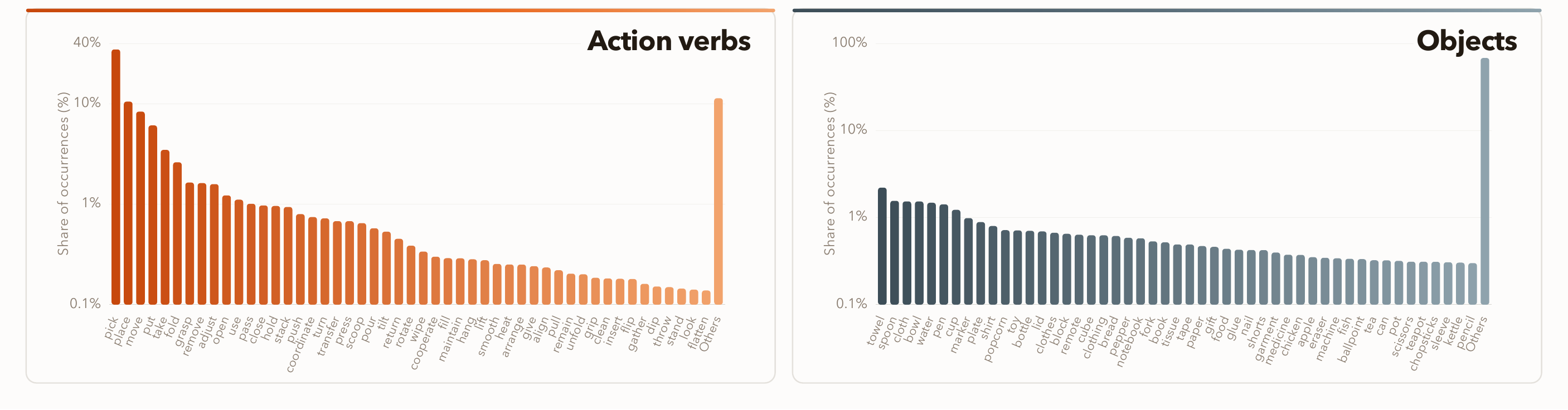}
\caption{\textbf{Action and object concept distribution in the pre-training corpus.}
We show the most frequent action verbs and object nouns extracted from the pre-training data; each bar gives the concept's share of occurrences (\%) on a logarithmic axis, and the trailing \emph{Others} bar aggregates the remaining vocabulary.}
\label{fig:data_analyze}
\end{figure}

To characterize the semantic coverage of the pre-training corpus, we further analyze the frequency distribution of action and object concepts. As shown in Fig.~\ref{fig:data_analyze}, both action verbs and object nouns exhibit a clear long-tailed distribution. High-frequency actions are dominated by general manipulation primitives such as picking, placing, moving, and putting, while the object vocabulary is concentrated on common household and tabletop entities. This distribution indicates that the corpus provides broad coverage of everyday robot manipulation scenarios while retaining a diverse tail of less frequent skills and objects.

\paragraph{Autolabeling pipeline.}
To enrich the annotation signals available in large robot manipulation corpora,
we build an automated multimodal labeling pipeline that converts raw episodes
into multi-granularity semantic annotations, visual grounding annotations, and
action-trajectory annotations. For language annotation, we first apply
rule-based temporal segmentation to identify candidate action segments and
keyframes, and then query multimodal model APIs such as
Gemini 3~\citep{gemini3pro_modelcard} and Doubao Seed 2.0
Pro~\citep{bytedance2026seed2} to generate action hints, atomic task
descriptions, and episode-level instructions. These
multi-granularity language annotations allow us to train and query the policy
under different instruction granularities and to construct CoT pairs for
training intermediate reasoning. For visual
grounding, we combine multimodal foundation models followed by SAM3
tracking~\citep{carion2025sam3} to generate per-frame bounding boxes and
segmentation masks for task-relevant objects. Finally, for 2D end-effector
traces, we compute bimanual end-effector
positions from robot joint poses using forward kinematics and project the
resulting 3D trajectories onto the head-camera image plane.

\paragraph{Web and VQA co-training.}
To retain the VLM's general language capability and broad generalization while
strengthening its spatial perception, we co-train with a large-scale
vision--language mixture spanning generic web VQA~\citep{liu2024llavanext,li2024llavaonevision},
embodied VQA~\citep{yuan2024robopoint,lee2025molmoact,ji2025robobrain},
and in-house annotations generated by the autolabeling pipeline above. The
in-house portion covers subtask decomposition, object bounding boxes, and
general commonsense VQA over our robot scenes.
During pre-training, VQA and action samples are combined in an action-heavy
mixture. Both sample types are optimized with the same next-token
cross-entropy loss over their target tokens, so language answers, CoT traces,
and action codes are all supervised within the unified autoregressive decoder.

\paragraph{Chain-of-thought supervision.}
Each robot sample is assigned exactly one CoT format, drawn by weighted random
sampling from eight candidates: a no-CoT baseline, atomic-task and high-level-task
text, subtask text, subtask-with-action-hint, 2D trajectory traces, and
bounding-box (object-localization) variants.
The subtask-text format is assigned a higher sampling weight, while the remaining formats use the default weighting. This mirrors the increased emphasis placed on the in-domain subtask-prediction split on the VLM side, reflecting a consistent focus on subtask grounding across both modalities.
Evaluation uses the fixed no-CoT format.

\paragraph{Implementation Details.}
\label{sec:pretrain-impl}
We optimize a single cross-entropy objective over the shared vocabulary using AdamW with a warmup phase followed by a constant phase and a final cosine decay, while keeping the vision tower unfrozen throughout.
The observation input provides a sparse multi-second history that includes the current frame. Historical frames are randomly dropped during training as regularization for the visual-memory module (Sec.~\ref{sec:visual_memory}), and the model is trained until convergence.

\section{Experiments}

We design our experiments to comprehensively probe the capabilities of \Gm along the axes that matter most for a general-purpose VLA: out-of-the-box deployability, transferability via fine-tuning, scalability to long-horizon tasks, fidelity to language, and adaptability to different contexts.
Concretely, our evaluation is organized around the following research questions:

\begin{itemize}[leftmargin=*]
    \item \textbf{Q1: How well does \Gm generalize beyond its training environments and objects?}
    After post-training on DROID data, we deploy \Gm on a DROID Franka platform whose physical environment and object instances are absent from both pre-training and post-training; we also stress-test its instruction-following capability under our Pick-and-Place Benchmark on the \RoneLite robot (Sec.~\ref{subsec:droid_zeroshot}, Sec.~\ref{subsec:pp_bench}).

    \item \textbf{Q2: How effectively can \Gm be adapted to out-of-domain benchmarks?}
    We fine-tune \Gm on external robot datasets, including DROID and Bridge, and evaluate it on the corresponding hardware and simulation suites. This measures how well the pretrained representation transfers when both the embodiment and the data distribution differ from our in-house platforms (Sec.~\ref{subsec:droid_zeroshot}, Sec.~\ref{subsec:bridge_simpler}).

    \item \textbf{Q3: How does \Gm perform on in-domain tasks after fine-tuning?}
    We evaluate two complementary in-domain settings: standardized simulation benchmarks (LIBERO, RoboTwin~2.0) for reproducibility and comparison with prior work, and real-world fine-tuning on \RoneLite and \RonePro across six task-embodiment settings to measure long-horizon bimanual manipulation under matched training and evaluation conditions (Sec.~\ref{subsec:libero}, Sec.~\ref{subsec:robotwin}, Sec.~\ref{subsec:real_world}).

    \item \textbf{Q4: Can \Gm acquire long-horizon generalist mobile manipulation skills, and how do architectural choices and pre-training data distribution shape this capability?}
    We evaluate \Gm with a single policy on the 2025 BEHAVIOR Challenge, a 50-task household benchmark where each episode averages 6.6 minutes and demands coordinated navigation and bimanual manipulation. We analyze how architectural choices and pre-training data distribution shape downstream long-horizon performance (Sec.~\ref{subsec:long_horizon_tasks}).

    \item \textbf{Q5: How strong is \Gm's language-following ability in cluttered scenes?}
    We introduce the Pick-and-Place Benchmark (PP Bench), which disentangles language grounding from low-level execution by separately reporting language following rate and task success rate across in-distribution and out-of-distribution object categories at multiple post-training scales (Sec.~\ref{subsec:pp_bench}).

    \item \textbf{Q6: How do different contexts affect \Gm's behavior?}
    We study how augmenting the policy input with additional referring context, such as cropped object/container regions and coordinate tokens provided by an external VLM, influences language grounding and final task success, leveraging the flexible multi-image interface of \Gm (Sec.~\ref{subsec:pp_bench}).

    \item \textbf{Q7: Does putting reasoning in the same stream as action actually pay off?}
    On a single pretrained checkpoint we toggle the action head (AR tokens vs.\ an additional flow-matching head) and the CoT stream (on/off) at inference time, across PP Bench and two new zero-shot long-horizon household tasks. We also qualitatively observe how per-stage instruction wording---e.g., adverbial qualifiers, spatial cues, or near-synonymous verbs---affects rollouts under AR$+$CoT (Sec.~\ref{subsec:cot_head}).
\end{itemize}

The remainder of this section is organized to answer each of these questions in turn.

\subsection{DROID Environment- and Object-Level Zero-Shot Evaluation}
\label{subsec:droid_zeroshot}

We first post-train \Gm on the DROID dataset~\citep{khazatsky2024droid} and then deploy it on a held-out DROID robot setup. No demonstrations from the physical evaluation environment or involving the evaluation object instances are included in either pre-training or post-training.
This evaluation therefore measures environment- and object-level zero-shot generalization after DROID post-training, rather than dataset-level zero-shot transfer.

\subsubsection{Evaluation Setup}

\paragraph{Robot Platform.}
We use a Franka Research 3 7-DoF robot arm equipped with a Robotiq 2F-85 parallel-jaw gripper, mounted on a height-adjustable standing desk following the standard DROID hardware configuration~\citep{khazatsky2024droid}.
Visual observations are provided by two RGB cameras: a right-side third-person camera offering a fixed global view of the tabletop workspace, and a wrist-mounted camera providing a close-up view for fine-grained manipulation.
The policy receives both camera streams together with the natural language task instruction.

\begin{figure*}[t]
    \centering

    \begin{subfigure}[t]{0.23\textwidth}
        \centering
        \includegraphics[width=\linewidth]{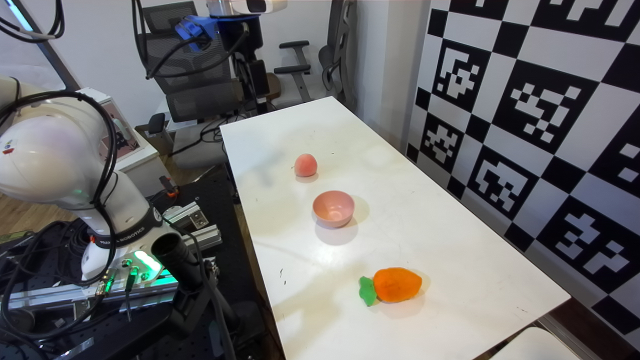}
        \caption{{\scriptsize Carrot / Peach $\rightarrow$ Bowl}}
    \end{subfigure}
    \hfill
    \begin{subfigure}[t]{0.23\textwidth}
        \centering
        \includegraphics[width=\linewidth]{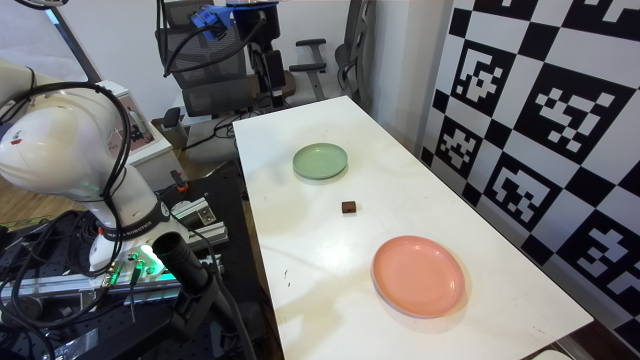}
        \caption{{\scriptsize Block $\rightarrow$ Green / Red Plate}}
    \end{subfigure}
    \hfill
    \begin{subfigure}[t]{0.23\textwidth}
        \centering
        \includegraphics[width=\linewidth]{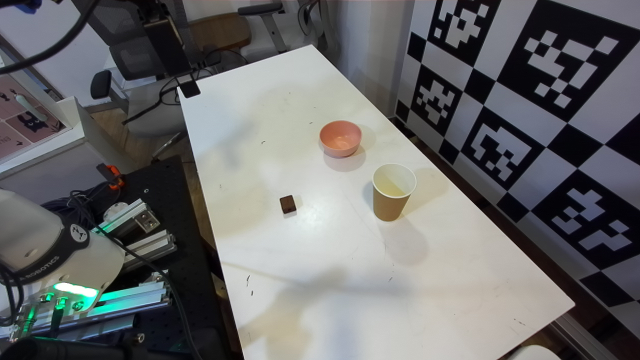}
        \caption{{\scriptsize Block $\rightarrow$ Cup}}
    \end{subfigure}
    \hfill
    \begin{subfigure}[t]{0.23\textwidth}
        \centering
        \includegraphics[width=\linewidth]{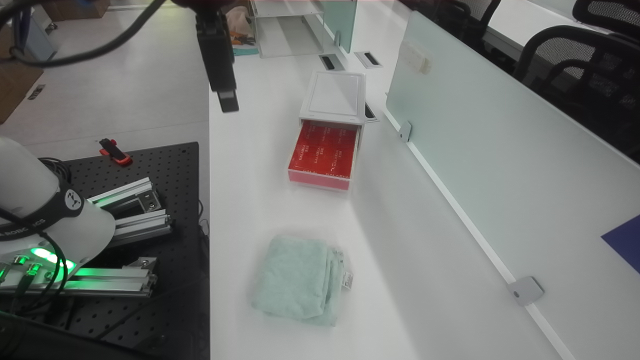}
        \caption{{\scriptsize Towel $\rightarrow$ Open Drawer}}
    \end{subfigure}

    \vspace{0.3em}

    \begin{subfigure}[t]{0.23\textwidth}
        \centering
        \includegraphics[width=\linewidth]{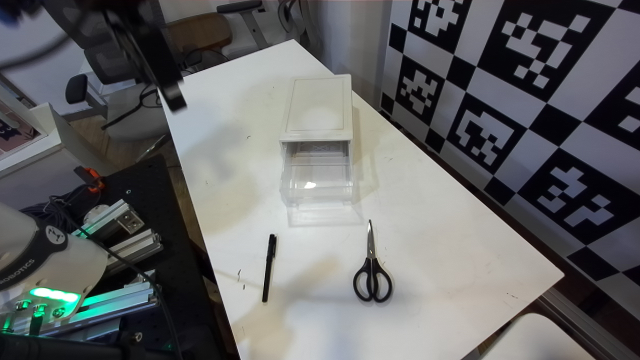}
        \caption{{\scriptsize Pen $\rightarrow$ Open Drawer}}
    \end{subfigure}
    \hfill
    \begin{subfigure}[t]{0.23\textwidth}
        \centering
        \includegraphics[width=\linewidth]{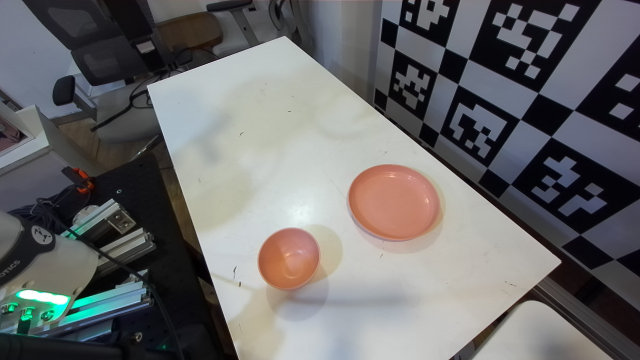}
        \caption{{\scriptsize Bowl $\rightarrow$ Left}}
    \end{subfigure}
    \hfill
    \begin{subfigure}[t]{0.23\textwidth}
        \centering
        \includegraphics[width=\linewidth]{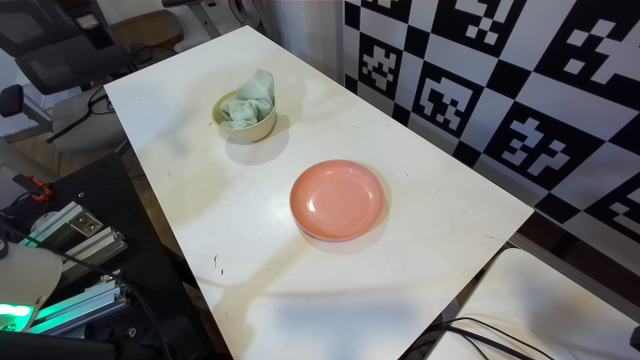}
        \caption{{\scriptsize Towel: Bowl $\rightarrow$ Plate}}
    \end{subfigure}
    \hfill
    \begin{subfigure}[t]{0.23\textwidth}
        \centering
        \includegraphics[width=\linewidth]{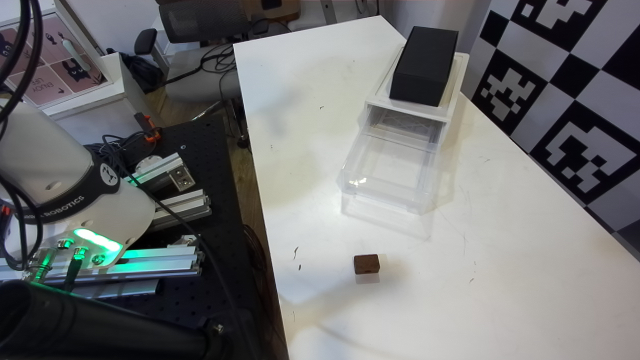}
        \caption{{\scriptsize Block $\rightarrow$ Drawer + Close}}
    \end{subfigure}

    \caption{\textbf{DROID environment- and object-level zero-shot evaluation tasks.} After DROID post-training, we evaluate \Gm on 10 manipulation tasks across 8 held-out scene setups and previously unseen physical object instances on a Franka Research 3 robot arm. Tasks cover object placement, color-conditioned target selection, small-aperture insertion, deformable object manipulation, spatial displacement, and multi-step sequential execution.}
    \label{fig:droid_tasks}
    \vspace{-0.5em}
\end{figure*}

\paragraph{Tasks.}
We evaluate on 10 tabletop manipulation tasks drawn from the DROID environment~\citep{khazatsky2024droid}, as illustrated in Fig.~\ref{fig:droid_tasks}.
Tasks span seven skill categories, each targeting a distinct manipulation challenge:

\begin{itemize}[leftmargin=*]
    \item \textbf{Move the carrot / peach into the bowl.} Requires discriminating between a soft deformable carrot plush and a rigid spherical peach, and precisely depositing the target object into a small bowl.
    \item \textbf{Move the block onto the green / red plate.} Requires colour-conditioned target selection and fine-grained grasping of a small block onto the correct plate.
    \item \textbf{Move the block into the cup.} Requires accurate vertical clearance estimation to deposit the block inside a tall cup without colliding with the rim.
    \item \textbf{Put the towel / pen into the open drawer.} Requires identifying the target object, localising the drawer opening, and grasping both deformable fabric and a thin rigid tool.
    \item \textbf{Move the bowl to the left.} Requires spatial-direction understanding and stable grasping of a wide, irregularly shaped bowl for precise lateral displacement.
    \item \textbf{Take out the towel from the bowl and put it on the plate.} A two-step sequential task: extract a deformable towel from a bowl, then reposition it onto a flat plate.
    \item \textbf{Put the block into the open drawer and close the drawer.} A long-horizon task requiring two temporally dependent sub-goals: block insertion followed by drawer closure.
\end{itemize}

\paragraph{Evaluation Protocol.}
Each task is evaluated over 10 trials.
Task success is scored as a binary outcome (1 for completion, 0 otherwise), except for the sequential task \textit{put the block into the open drawer and close the drawer}, which receives a partial score of 0.5 for completing only the insertion sub-step and a full score of 1.0 for full task completion.

\paragraph{Baselines.}
We compare \Gm against two representative baselines: $\pi_{0.5}$-DROID~\citep{pi0.5}, trained on the original DROID dataset with a PaliGemma backbone, and MolmoAct2-DROID~\citep{fang2026molmoact2actionreasoningmodels}, a generalist policy built on the Molmo vision-language model and trained with the MolmoAct2 data preprocessing pipeline.

\subsubsection{Quantitative Results}

\begin{figure}[thpb]
    \centering
    \includegraphics[width=0.95\columnwidth]{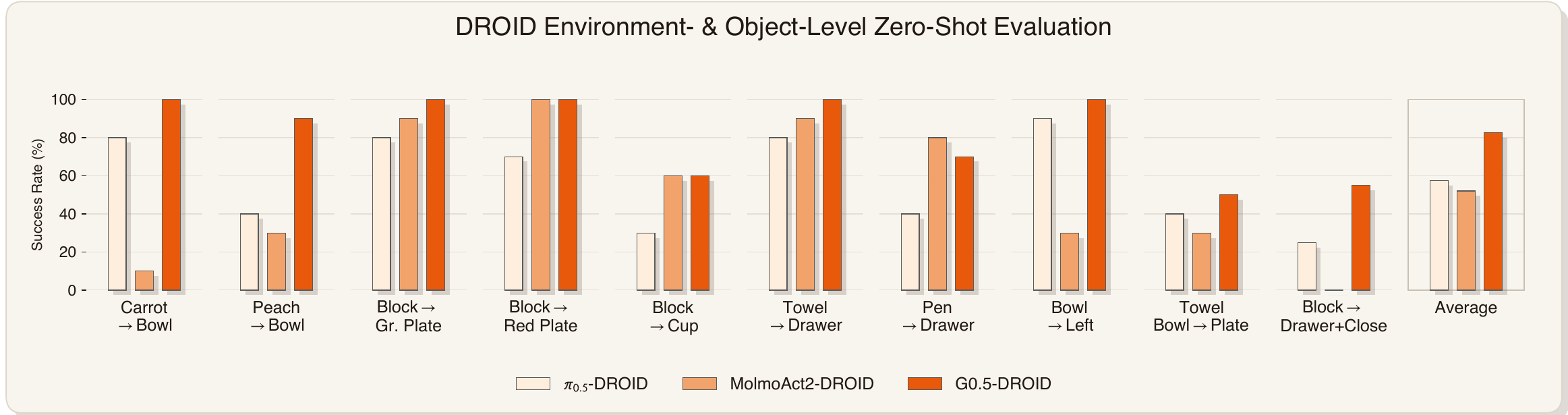}
    \caption{\textbf{DROID environment- and object-level zero-shot evaluation results.} All models are trained on DROID data, while the physical evaluation environment and object instances are held out. Per-task success rates (\%) are reported for $\pi_{0.5}$-DROID, MolmoAct2-DROID, and \Gm across 10 manipulation tasks. \Gm achieves an average of 82.5\%, outperforming $\pi_{0.5}$-DROID (57.5\%) by 25.0~percentage points and MolmoAct2-DROID (52.0\%) by 30.5~percentage points.}
    \label{fig:droid_results}
\end{figure}

Fig.~\ref{fig:droid_results} presents the per-task success rates across all three models.
Overall, \Gm consistently outperforms both baselines across the majority of tasks, achieving an average success rate of 82.5\%.

\paragraph{\Gm-DROID vs.\ $\pi_{0.5}$-DROID.}
\Gm-DROID outperforms $\pi_{0.5}$-DROID on all 10 tasks, with particularly strong advantages on tasks that demand precise object discrimination and multi-step reasoning.
On tasks where objects share similar appearance or require colour-conditioned target selection, \Gm-DROID demonstrates significantly stronger visual grounding.

\paragraph{\Gm-DROID vs.\ MolmoAct2-DROID.}
\Gm-DROID shows especially large margins on tasks involving spatial language instructions and object-category recognition, where MolmoAct2 struggles to ground instruction semantics into correct motor behaviour.
Most notably, MolmoAct2-DROID completely fails on the sequential task \textit{put the block into the open drawer and close the drawer}, while \Gm-DROID succeeds on over half of the trials, demonstrating substantially stronger multi-stage task execution capability.
We further observe that MolmoAct2-DROID frequently freezes or produces no motion when approaching objects such as the carrot, peach, or bowl, and often executes empty grasps when the gripper has not yet reached a valid pre-grasp pose.

\paragraph{Effect of visual contrast on drawer localisation.}
The drawer cabinet used in this evaluation features a white semi-transparent body, which provides limited visual contrast for localising the insertion aperture.
To examine the impact of this visual ambiguity, we conducted a controlled comparison on the towel insertion task: in the main evaluation, orange adhesive cards were attached to the drawer's interior walls and base as explicit localisation markers, whereas the earlier experiment was run without any markers ($\pi_{0.5}$-DROID: 90\%, MolmoAct2-DROID: 80\%, \Gm-DROID: 60\%).
Adding the high-contrast markers dramatically improves \Gm-DROID's performance to 100\%, while $\pi_{0.5}$-DROID remains comparatively unaffected.
This indicates that \Gm-DROID is more sensitive to low-contrast semi-transparent surfaces and is relatively less capable of reliably localising targets without explicit high-contrast visual cues.

\subsection{Simulation Benchmarks}
\subsubsection{Bridge-SimplerEnv}
\label{subsec:bridge_simpler}

Bridge-SimplerEnv evaluates language-conditioned WidowX manipulation policies in SimplerEnv, a real-to-sim benchmark that instantiates the BridgeData V2/WidowX setup for scalable simulated policy evaluation \citep{simplerenv}. We follow the official SimplerEnv WidowX+Bridge task suite and evaluate on four Bridge-style manipulation tasks: putting a spoon on a towel, putting a carrot on a plate, stacking a green cube on a yellow cube, and putting an eggplant into a yellow basket. To adapt our policy to the WidowX/Bridge embodiment before evaluation, we post-train it on BridgeData V2 demonstrations \citep{walke2023bridgedatav2} for 80K gradient steps with a learning rate of $3\times10^{-5}$. Throughout both post-training and evaluation, we adopt a state-free policy input, where robot joint states or other proprioceptive states are not provided to the model. This yields a stricter yet comparable evaluation protocol under the same SimplerEnv task suite. We then evaluate the resulting policy in SimplerEnv without any additional simulation-domain training. Results are summarized in Tab.~\ref{tab:simplerenv_bridge}, where \Gm achieves the highest average success rate of $87.3\%$ among the compared methods.

\begin{table*}[t]
\centering
\small
\caption{Results on Bridge-SimplerEnv. All numbers are success rates (\%). Since several original model papers do not report results on SimplerEnv-WidowX, we compile the corresponding baseline numbers from prior studies that evaluate these models on this benchmark. The best and second-best results are highlighted in bold and underlined, respectively.}
\label{tab:simplerenv_bridge}
\setlength{\tabcolsep}{3.5pt}
\renewcommand{\arraystretch}{1.12}
\begin{tabular}{lccccc}
\toprule
Method
& \makecell{Put Spoon\\on Towel}
& \makecell{Put Carrot\\on Plate}
& \makecell{Stack Green Block\\on Yellow Block}
& \makecell{Put Eggplant\\in Yellow Basket}
& \textbf{Average} \\
\midrule

$\pi_0$ \citep{pi0}
& 29.1
& 0.0
& 16.6
& 62.5
& 27.1 \\

$\pi_0$-FAST \citep{fast}
& 29.1
& 21.9
& 10.8
& 66.6
& 32.1 \\

$\pi_{0.5}$ \citep{pi0.5}
& 49.3
& \underline{64.7}
& 44.7
& 69.7
& 57.1 \\

GR00T-N1.5 \citep{gr00t-n1}
& 75.3
& 54.3
& 57.0
& 61.3
& 61.9 \\

StarVLA-GR00T \citep{starvla}
& 83.0
& 59.4
& 18.8
& \best{100.0}
& 65.3 \\

RoboBrain2.5-8B \citep{ji2025robobrain}
& 75.0
& 55.5
& 40.1
& \best{100.0}
& 67.6 \\

MemoryVLA \citep{memoryvla}
& 75.0
& \best{75.0}
& 37.5
& \best{100.0}
& 71.9 \\

EO-1 \citep{eo1}
& 63.6
& 54.5
& \underline{81.8}
& 90.9
& 72.7 \\

Xiaomi-Robotics-0 \citep{xiaomirobotics0}
& \underline{95.8}
& 62.5
& 75.0
& 83.3
& \underline{79.2} \\


\midrule
\textbf{\Gm (Ours)}
& \best{97.5}
& \best{75.0}
& \best{83.3}
& \underline{93.3}
& \best{87.3} \\

\bottomrule
\end{tabular}
\end{table*}

\subsubsection{RoboTwin 2.0}
\label{subsec:robotwin}
RoboTwin 2.0 evaluates simulated bimanual manipulation across a broad suite of
over 50 tasks, emphasizing behaviors that depend on coordinated dual-arm
control rather than single-arm pick-and-place skills alone. We follow the
multi-task training setup of~\citep{motus, lingbotva}:
models are trained on a combined set of 2,500 clean-scene demonstrations and
25,000 demonstrations collected with heavy scene randomization. We finetune
\Gm for 4 epochs using a learning rate of $4\times10^{-5}$ and a global batch
size of 1024. Success rates are averaged over 100 trials per task in both clean
and randomized evaluation settings. Results are reported in
Tab.~\ref{tab:robotwin}.
Per-task success rates for \Gm are provided in Tab.~\ref{tab:robotwin_detail}.

\begin{table}[t]
\centering
\small
\caption{Results on RoboTwin 2.0. We report success rates under clean and
randomized evaluation settings, together with their average. The best and second-best results are highlighted in bold and underlined,
respectively.}
\label{tab:robotwin}
\begin{tabular}{lccc}
\toprule
Method & Clean & Rand. & \textbf{Average} \\
\midrule
$\pi_{0}$~\citep{pi0} & 65.9 & 58.4 & 62.2 \\
$\pi_{0.5}$~\citep{pi0.5} & 82.7 & 76.8 & 79.8 \\
LingBot-VLA~\citep{lingbotvla2026} & 86.5 & 85.3 & 85.9 \\
Motus~\citep{motus} & 88.7 & 87.0 & 87.8 \\
StarVLA~\citep{starvla} & 88.7 & 87.8 & 88.3 \\
LingBot-VA~\citep{lingbotva} & \underline{92.9} & 91.5 & \underline{92.2} \\
Fast-WAM~\citep{fastwam} & 91.9 & \underline{91.8} & 91.8 \\
\midrule
\textbf{\Gm (Ours)} & \best{93.7} & \best{92.8} & \best{93.3} \\
\bottomrule
\end{tabular}
\end{table}

\subsubsection{LIBERO} \label{subsec:libero}

LIBERO is a Franka robot arm simulation benchmark comprising four task suites—Goal, Spatial, Object, and Long—which evaluate instruction following, spatial reasoning, object recognition, and long-horizon manipulation, respectively. Each suite contains 10 tasks with 50 demonstrations per task. We finetune \Gm for 100K steps using a learning rate of $1\times10^{-5}$ and a weight decay of $1\times10^{-2}$. Following the standard LIBERO protocol, we evaluate the model over 50 rollout trials per task using the benchmark-defined initial states. Results are summarized in Tab.~\ref{tab:libero}. \Gm achieves state-of-the-art overall performance among recent VLA models, attaining an average success rate of $98.9\%$. Notably, it delivers the strongest performance on the challenging Long suite.

\begin{table*}[t]
\centering
\small
\caption{Results on LIBERO. The best and second-best results are highlighted in bold and underlined, respectively. }
\label{tab:libero}
\begin{tabular}{lccccc}
\toprule
Method & Spatial & Object & Goal & Long & \textbf{Average} \\
\midrule

Qwen-VLA-Base \citep{qwenvla}
& -- & -- & -- & -- & 90.8 \\

$\pi_0$ \citep{pi0}
& 98.0 & 96.8 & 94.4 & 88.4 & 94.4 \\

InternVLA-M1 \citep{internvlam1}
& 98.0 & 99.0 & 93.8 & 92.6 & 95.9 \\

Wall-OSS-0.5 \citep{walloss}
& -- & -- & -- & -- & 96.5 \\

$\pi_{0.5}$ \citep{pi0.5}
& 98.8 & 98.2 & 98.0 & 92.4 & 96.9 \\

GR00T-N1.7 \citep{gr00t-n1}
& 97.7 & 98.5 & 97.5 & 94.4 & 97.0 \\

OpenVLA-OFT \citep{openvlaoft}
& 97.6 & 98.4 & 97.9 & 94.5 & 97.1 \\

Fast-WAM \citep{fastwam}
& 98.2 & \best{100.0} & 97.0 & 95.2 & 97.6 \\

Being-H0.5 \citep{being-h05}
& \underline{99.2} & 98.2 & \underline{99.0} & 96.2 & 98.2 \\

Motus \citep{motus}
& 96.8 & \underline{99.8} & 96.6 & 97.6 & 97.7 \\

Qwen-VLA-Instruct \citep{qwenvla}
& -- & -- & -- & -- & 97.9 \\


EO1 \citep{eo1}
& \best{99.7}
& \underline{99.8}
& \best{99.2}
& 94.8
& 98.4 \\

Cosmos Policy \citep{cosmospolicy}
& 98.1 & \best{100.0} & 98.2 & 97.6 & 98.5 \\

LingBot-VA \citep{lingbotva}
& 98.5 & 99.6 & 97.2 & \underline{98.5} & 98.5 \\

Xiaomi-Robotics-0 \citep{xiaomirobotics0}
& 98.8 & \best{100.0} & 98.8 & 97.2 & \underline{98.7} \\

\midrule
\textbf{\Gm (Ours)}
& 98.4
& \best{100.0}
& 98.6
& \best{98.6}
& \best{98.9} \\

\bottomrule
\end{tabular}
\end{table*}

\subsection{Long-horizon Tasks}
\label{subsec:long_horizon_tasks}

The 2025 BEHAVIOR Challenge, built on the BEHAVIOR-1K benchmark~\citep{li2023behavior} and the photo-realistic OmniGibson simulator powered by NVIDIA Isaac Sim, presents a demanding testbed for \textbf{long-horizon mobile manipulation}. The challenge selects 50 full-length household tasks from the 1,000 activity collection, covering diverse activities like rearrangement, cooking, cleaning, and installation. To support training, it provides 10,000 teleoperated expert demonstrations totaling over 1,100 hours, where each demonstration episode averages 6.6 minutes and spans up to 14 minutes. To accomplish these tasks, a policy must control an \textbf{\RonePro} robot to simultaneously process RGB observations from the head and dual-wrist cameras, navigate through house-scale environments, and perform dexterous bimanual manipulation using two 7-DOF arms equipped with parallel-jaw grippers.

During evaluation, policies are tested over 10 episodes per task with systematically randomized initial object states and robot poses.
Because executing these complex household chores is significantly more demanding than short horizon table top tasks, overall performance is quantified by a Task Success Score, which serves as the primary ranking metric for the challenge. This metric measures how much of a goal condition a policy satisfies by computing the proportion of completed BDDL goal predicates and selecting the best matched goal clause at the end of the episode. By awarding partial credit, it ensures that policies making meaningful progress score higher even without full task completion. Consequently, this granular scoring mechanism provides a smoother and more reliable way to evaluate incremental progress and compare policies across BEHAVIOR tasks than a traditional binary success rate.

\subsubsection{Implementation Details}

For our evaluation, we adopt the official standard track and the default low-resolution RGB rendering setting. 
To ensure a fair comparison, we follow the first place solution~\citep{1st_place} and use single frame observations during post training, but omit their explicit stage head.
Notably, to validate the general mobile manipulation capabilities of our pre-trained model and to measure the comprehensive performance of a single policy across 50 diverse household tasks, we jointly co-train all 10,000 episodes from the 50 tasks during the post-training phase. Consequently, the test results reported below for both $\pi_{0.5}$ and \Gm are evaluated using only a single checkpoint.

\begin{table}[h!]
\centering
\small
\caption{Overall results on the 2025 BEHAVIOR Challenge (50 tasks, 10 instances each). \textit{Task Success Score} is the challenge ranking metric (task progress). The first place solution by the Robot Learning Collective (RLC)~\citep{1st_place} uses a set of 4 checkpoints; $\pi_{0.5}$~\citep{pi0.5} (4 epochs) and \Gm each use a single checkpoint, averaged over two eval runs. Best/second best in \textbf{bold}/\underline{underline}.}
\vspace{2mm}
\label{tab:b1k_overall_results}
\begin{tabular}{lc}
\toprule
\textbf{Method} & \textbf{Task Success Score (Ranking Metric)} $\uparrow$ \\
\midrule
RLC (1st place)~\citep{1st_place} & 0.2605 \\
Comet (2nd place)~\citep{comet} & 0.1830 \\
$\pi_{0.5}$ (4 epochs) ~\citep{pi0.5} & 0.2626 \\
\textbf{\Gm (Ours, 1 epoch)} & \underline{0.2904} \\
\textbf{\Gm (Ours, 4 epochs)} & \best{0.3136} \\
\bottomrule
\end{tabular}
\end{table}

\subsubsection{Key Findings and Analysis}

We evaluate \Gm and $\pi_{0.5}$ directly to establish a strictly fair comparison of the pre-trained model weights using a single policy, whereas other baseline scores represent public leaderboard submissions utilizing multiple policies. From \Cref{tab:b1k_overall_results} and the detailed results in \Cref{tab:b1k_results}, we highlight the following key findings:

\begin{itemize}[leftmargin=*]
    \item \textbf{Training Efficiency.}
    With only a single epoch of post-training, \Gm already surpasses $\pi_{0.5}$ trained for four epochs by +10.6\% in the primary Task Success Score.
    With four epochs, this advantage widens to +19.4\%, demonstrating that our model continues to improve with additional training. 
    This result directly demonstrates the superior representational capacity of the \Gm pre-trained backbone: a stronger foundation model can extract task-relevant knowledge from the same downstream data far more efficiently, requiring significantly fewer gradient steps to acquire complex household manipulation skills.

    \item \textbf{Single-Policy Generalization.} Across the entire suite of 50 tasks, \Gm (4 epochs) outperforms the first-place solution by +20.4\% using only a single checkpoint, whereas the competition winner relies on a set of four distinct checkpoints to cover different task distributions. Even with just 1 epoch, \Gm already exceeds the first-place result by +11.5\%. This confirms that \Gm learns a sufficiently general whole-body control prior during pre-training, eliminating the need for task-specific checkpoint selection at evaluation time.
\end{itemize}

We attribute these results primarily to three aspects of \Gm's design:

\paragraph{Structured Action Decomposition Benefits Mobile Manipulation.}
As described in \Cref{sec:codec}, our structured tokenization decomposes the robot's action space into independent motion parts (e.g., left control, right control, lower body). This decomposition is particularly beneficial for mobile manipulation tasks, as it explicitly decouples navigation from manipulation in the token space. Rather than learning from a flat, entangled action representation, the model acquires a factored whole-body control prior where each motion group can be independently predicted. This advantage is clearly reflected in our results: \Gm achieves strong performance on long-horizon tasks that interleave navigation and object manipulation, such as \textit{moving boxes to storage} (+0.35 vs.\ $\pi_{0.5}$), \textit{picking up trash} (+0.30), and \textit{loading the car} (+0.28), where the robot must navigate to different locations, grasp objects, and place them at target positions in sequence.

\paragraph{Pre-Training Distribution Shapes Downstream Strengths.}
Our per-task analysis reveals a clear alignment between pre-training data composition and downstream task performance.  As shown in \Cref{fig:data_analyze}, the real-robot pre-training data for \Gm is predominantly composed of pick-and-place behaviors.
Correspondingly, \Gm demonstrates strong advantages on \textit{open-space pick-and-place} tasks, where the robot picks up objects and places them at target locations across diverse spatial configurations. Examples include \textit{setting mousetraps} (+0.46), \textit{assembling gift baskets} (+0.26), and \textit{putting shoes on rack} (+0.20), all of which primarily require robust grasping, accurate placement, and coordinated navigation across diverse spatial configurations.

Conversely, $\pi_{0.5}$ outperforms \Gm on \textit{container-interaction} tasks that involve appliance or cabinet manipulation (e.g., \textit{make microwave popcorn}: 0.95 vs.\ 0.55; \textit{cook hot dogs}: 0.93 vs.\ 0.90). These skills are severely underrepresented in our pre-training data; however, the gap narrows substantially with more training: \textit{cook hot dogs} improves from 0.45 (1 epoch) to 0.90 (4 epochs), approaching $\pi_{0.5}$'s 0.93.

Despite this distributional gap, \Gm leads on 29 out of 50 evaluated household tasks (58\%) while $\pi_{0.5}$ leads on only 15 (30\%), with 6 tasks being comparable. This broad coverage underscores the generality of the pre-trained representations. Moreover, it suggests a clear path forward: enriching the pre-training data with container-interaction skills could further narrow the remaining gap.

\paragraph{Visual Memory Pre-Training Improves Long-Horizon Performance.}
As described in \Cref{sec:visual_memory}, \Gm is pre-trained with factorized spatial-temporal attention that processes multi-frame visual context. Although we use single-frame input during post-training for fair comparison, the benefits of temporal pre-training are clearly evident in the downstream results. The gains are most pronounced on navigation-intensive, long-horizon tasks: \textit{moving boxes to storage}, \textit{loading the car}, \textit{bringing in wood}, and \textit{tidying bedroom}. These tasks require the robot to repeatedly traverse between distant locations while tracking which objects have been moved and where they were placed.

We attribute this advantage to the visual dynamics inherent in mobile manipulation pre-training data: even in standard pick-and-place episodes, the robot frequently moves its base between grasp and place locations, causing consecutive frames from each camera to exhibit large visual changes including scene layout shifts and object appearance transitions. The factorized temporal attention in the vision encoder, operating within each camera view across time steps is well-suited to capture these sequential visual dynamics, encouraging each per-camera representation to encode not just the current observation but also an implicit understanding of how the scene evolves over time. Additionally, during pre-training we stochastically drop all historical frames with 30\% probability to prevent the model from overfitting to historical context. Together, these design choices yield single-frame representations that are more spatially informed, which explains why \Gm generalizes well to long-horizon mobile manipulation even when post-trained with single-frame input.

\subsection{Real-World Fine-Tuning Evaluation}
\label{subsec:real_world}
We evaluate \Gm through real-world fine-tuning experiments on two robot embodiments, \RoneLite and \RonePro.
This evaluation focuses on whether a policy can be adapted to robot embodiments with different kinematic structures and execute long-horizon bimanual manipulation tasks under matched training and evaluation conditions.

\paragraph{Robot Embodiments.}

\RoneLite is a mobile dual-arm manipulation platform, comprising two 6-DoF arms, a 3-DoF torso, and a mobile omnidirectional base.
The torso provides vertical and pitching motion to extend the manipulation workspace and improve operational flexibility.

\RonePro is a humanoid upper-body mobile manipulation platform, comprising two 7-DoF arms, a 4-DoF torso, and a mobile omnidirectional base.
Compared with \RoneLite, the additional arm and torso degrees of freedom provide enhanced dexterity and greater flexibility for complex bimanual manipulation tasks.
Evaluating on both embodiments allows us to test whether the learned policy remains effective under different workspace, dexterity, and whole-body coordination requirements.

\paragraph{Baselines and Fine-Tuning Protocol.}
We compare \Gm with two representative open-source VLA baselines, $\pi_{0.5}$~\citep{pi0.5} and GR00T-N1.7~\citep{gr00t-n1}.
For each evaluation setting, all models are fine-tuned on the same training data with an aligned compute budget.
Specifically, each model is trained using 16 H20 GPUs for the same wall-clock duration within the same setting, ranging from 4 to 10 hours depending on task complexity.
All models are adapted to both robot embodiments and evaluated with the same observation space, action space, inference procedure, and low-level control settings.

\paragraph{Evaluation Protocol.}
We use \emph{task} to denote the semantic task objective, such as towel folding or carton folding, and \emph{setting} to denote a specific task-embodiment pair, such as towel folding on \RoneLite or towel folding on \RonePro.
Under this definition, the real-world fine-tuning evaluation contains four tasks instantiated as six evaluation settings.
The \RoneLite settings include towel folding, carton folding, and pencil-case packing, while the \RonePro settings include towel folding, carton folding, and box transfer and stacking.
Towel folding and carton folding are evaluated on both embodiments, enabling direct comparison across different robot configurations under the same task objectives.

Each setting is evaluated over 15 real-world episodes.
We report both task success rate and process score.
The success rate measures the fraction of episodes in which the full task is completed, while the process score evaluates intermediate task progress based on predefined stage-wise criteria.
The detailed stage definitions and scoring rules are provided in the appendix.
To reduce the influence of uncontrolled environmental factors, such as lighting changes and robot hardware state variations, we evaluate the models in an interleaved order within each setting rather than evaluating one model exhaustively before the next.

\paragraph{Observation Setup.}
Both platforms are equipped with three RGB cameras for visual observation.
Two wrist-mounted cameras provide close-up views for fine-grained gripper manipulation, while one external camera provides a global view of the scene for spatial understanding and long-horizon planning.

\subsubsection{Task Definitions and Evaluation Metrics}
\begin{figure*}[t]
    \centering

    \begin{subfigure}[t]{0.32\textwidth}
        \centering
        \includegraphics[width=\linewidth]{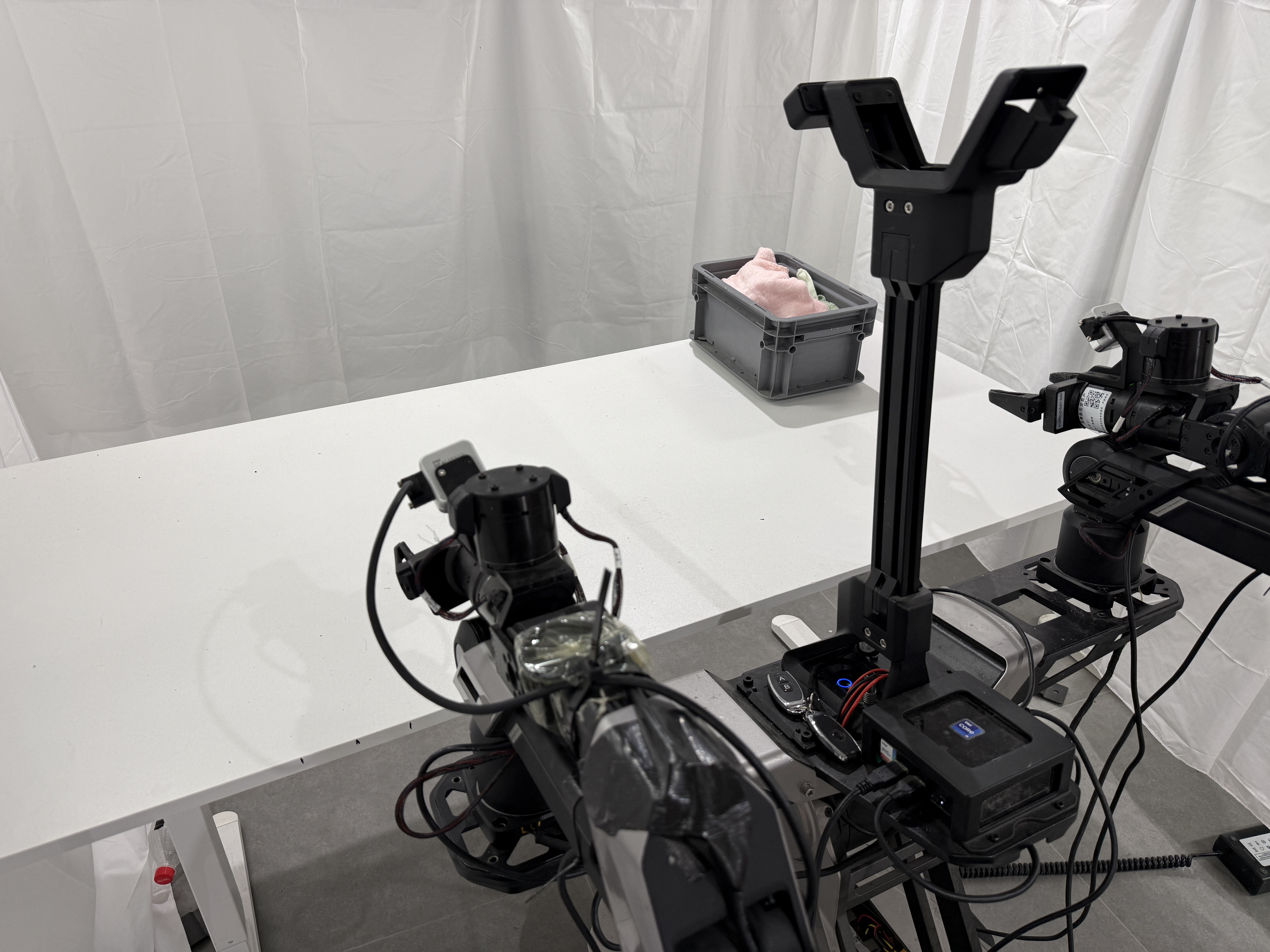}
        \caption{\RoneLite: Folding Towel}
    \end{subfigure}
    \hfill
    \begin{subfigure}[t]{0.32\textwidth}
        \centering
        \includegraphics[width=\linewidth]{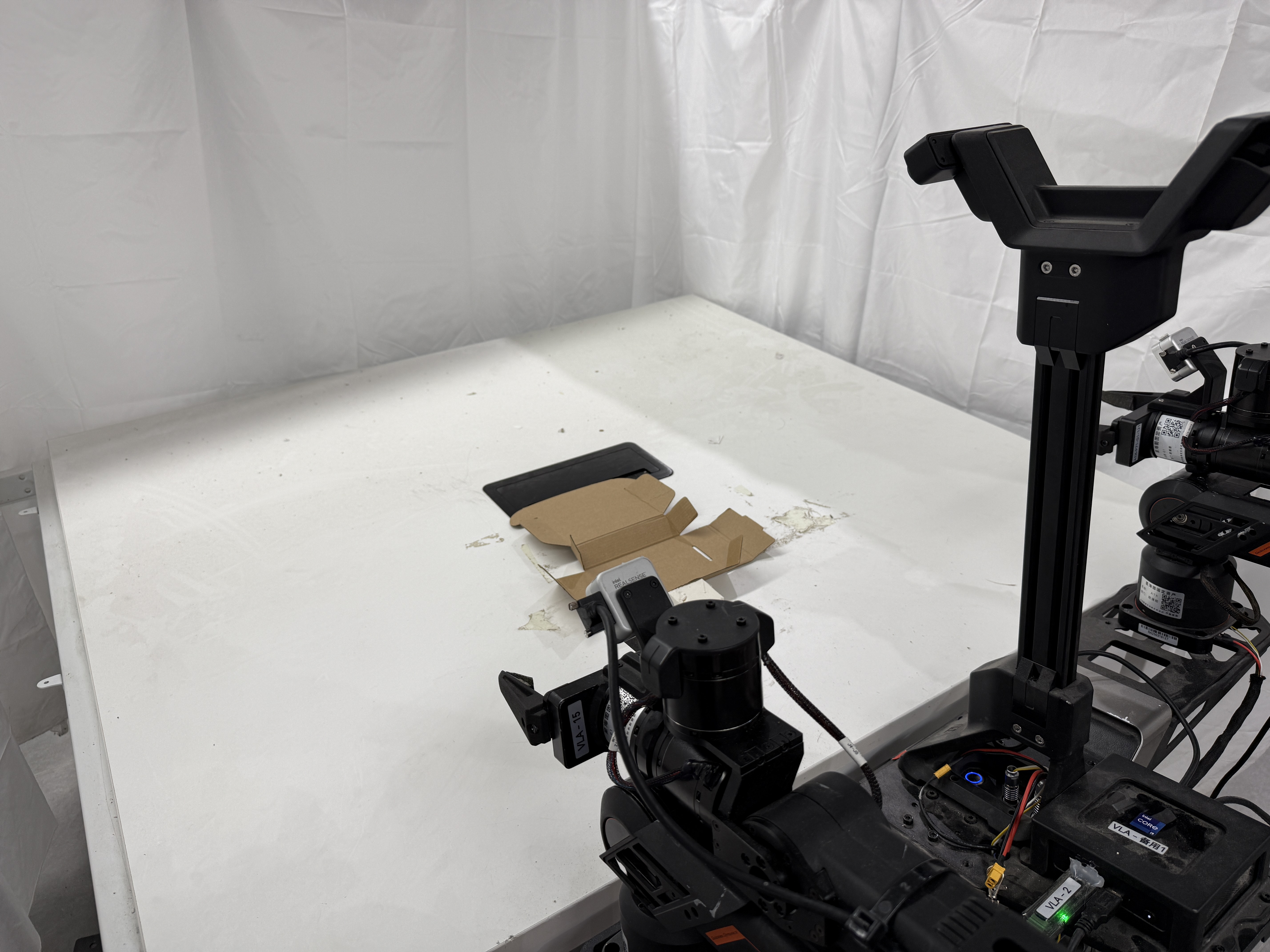}
        \caption{\RoneLite: Folding Carton}
    \end{subfigure}
    \hfill
    \begin{subfigure}[t]{0.32\textwidth}
        \centering
        \includegraphics[width=\linewidth]{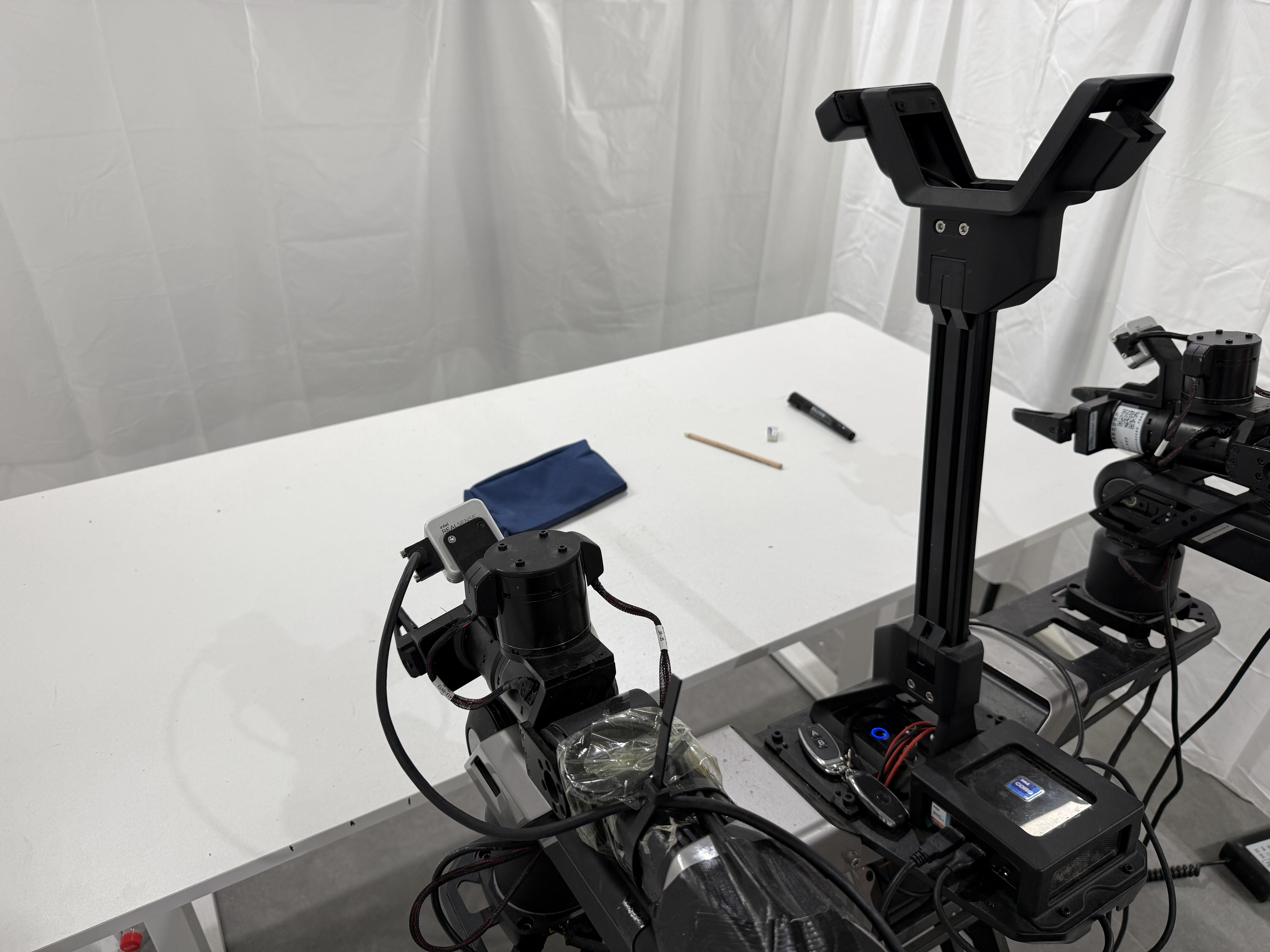}
        \caption{\RoneLite: Pencil-Case Packing}
    \end{subfigure}

    \vspace{0.3em}

    \begin{subfigure}[t]{0.32\textwidth}
        \centering
        \includegraphics[width=\linewidth]{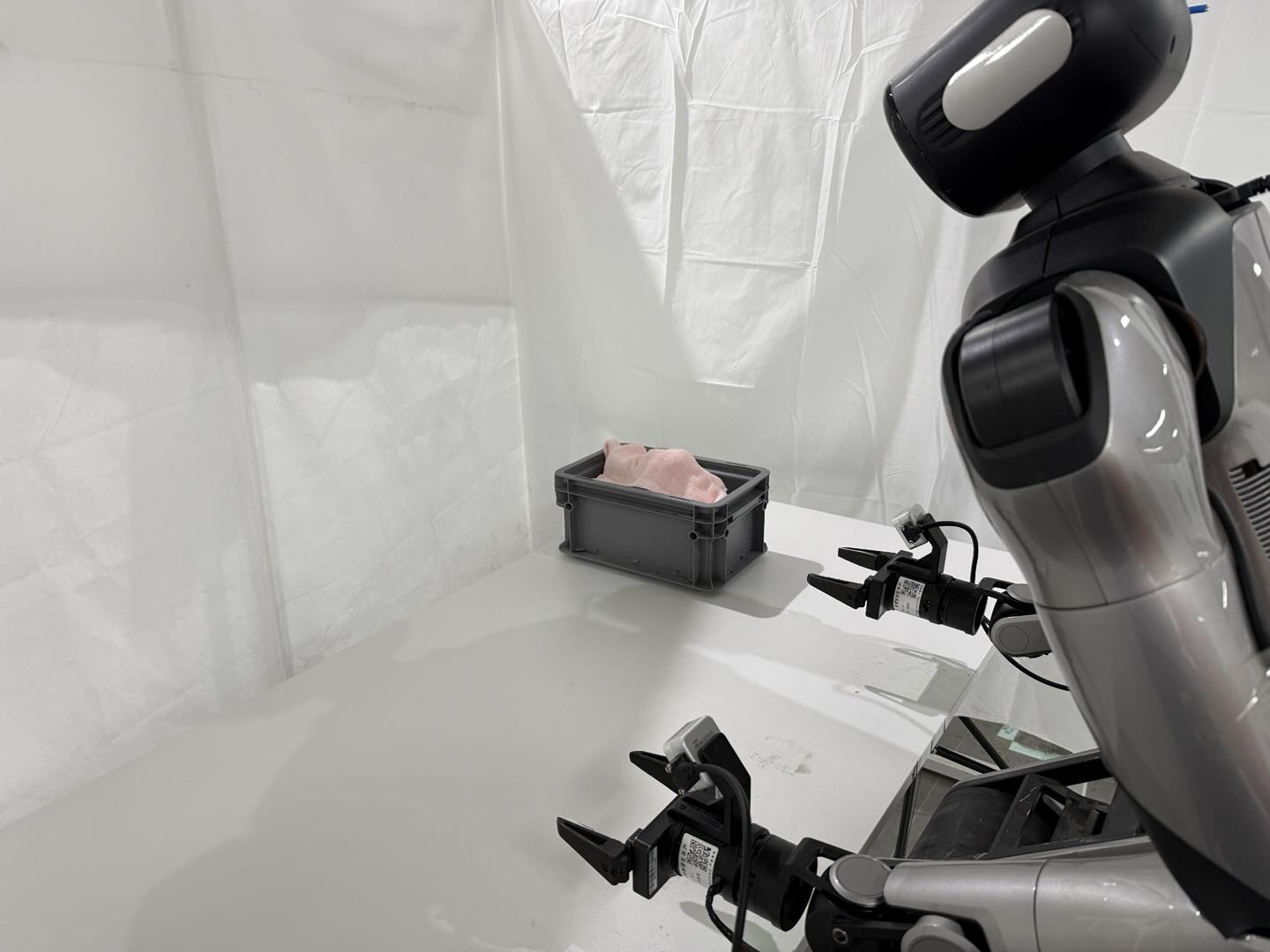}
        \caption{\RonePro: Folding Towel}
    \end{subfigure}
    \hfill
    \begin{subfigure}[t]{0.32\textwidth}
        \centering
        \includegraphics[width=\linewidth]{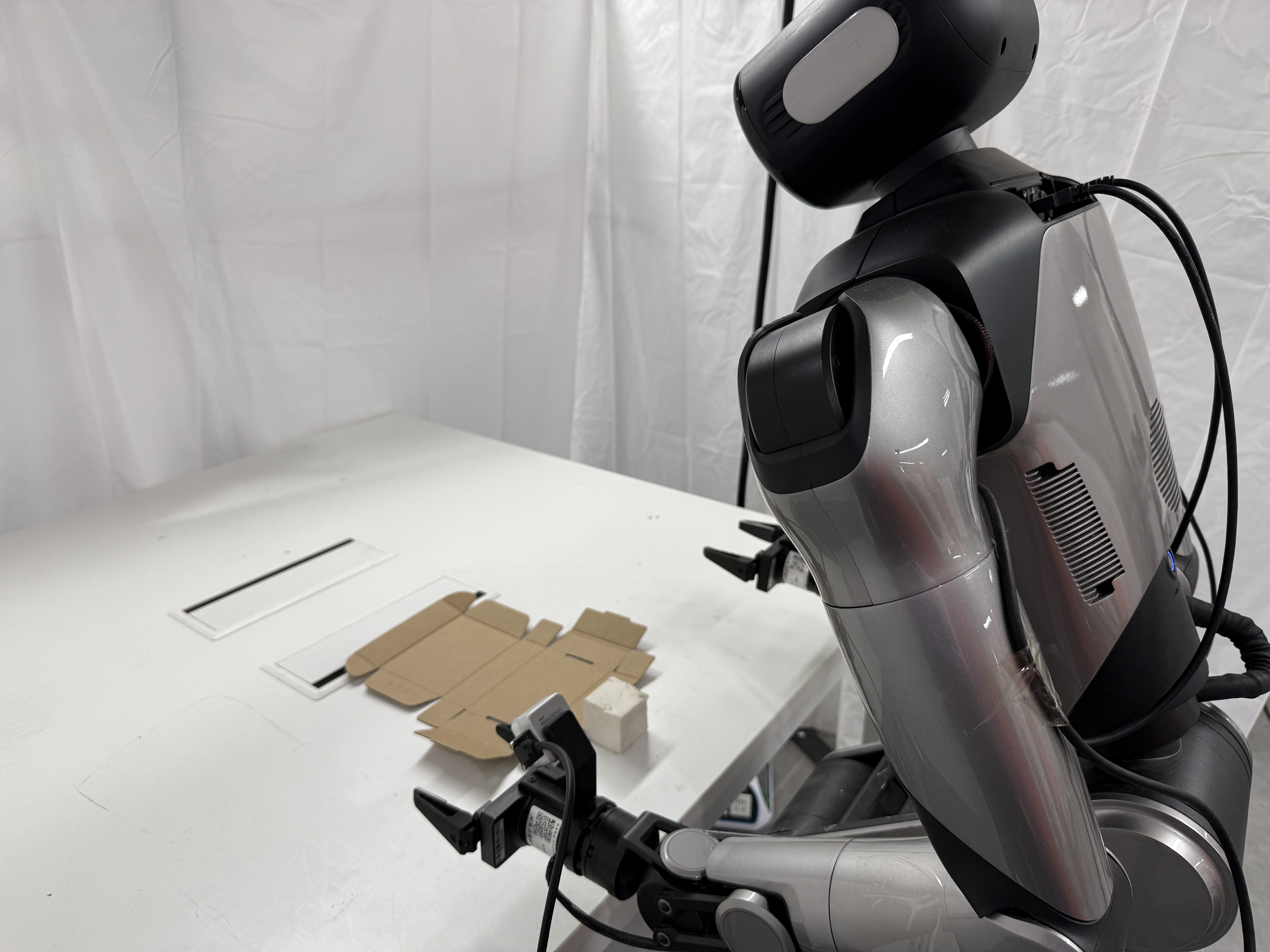}
        \caption{\RonePro: Folding Carton}
    \end{subfigure}
    \hfill
    \begin{subfigure}[t]{0.32\textwidth}
        \centering
        \includegraphics[width=\linewidth]{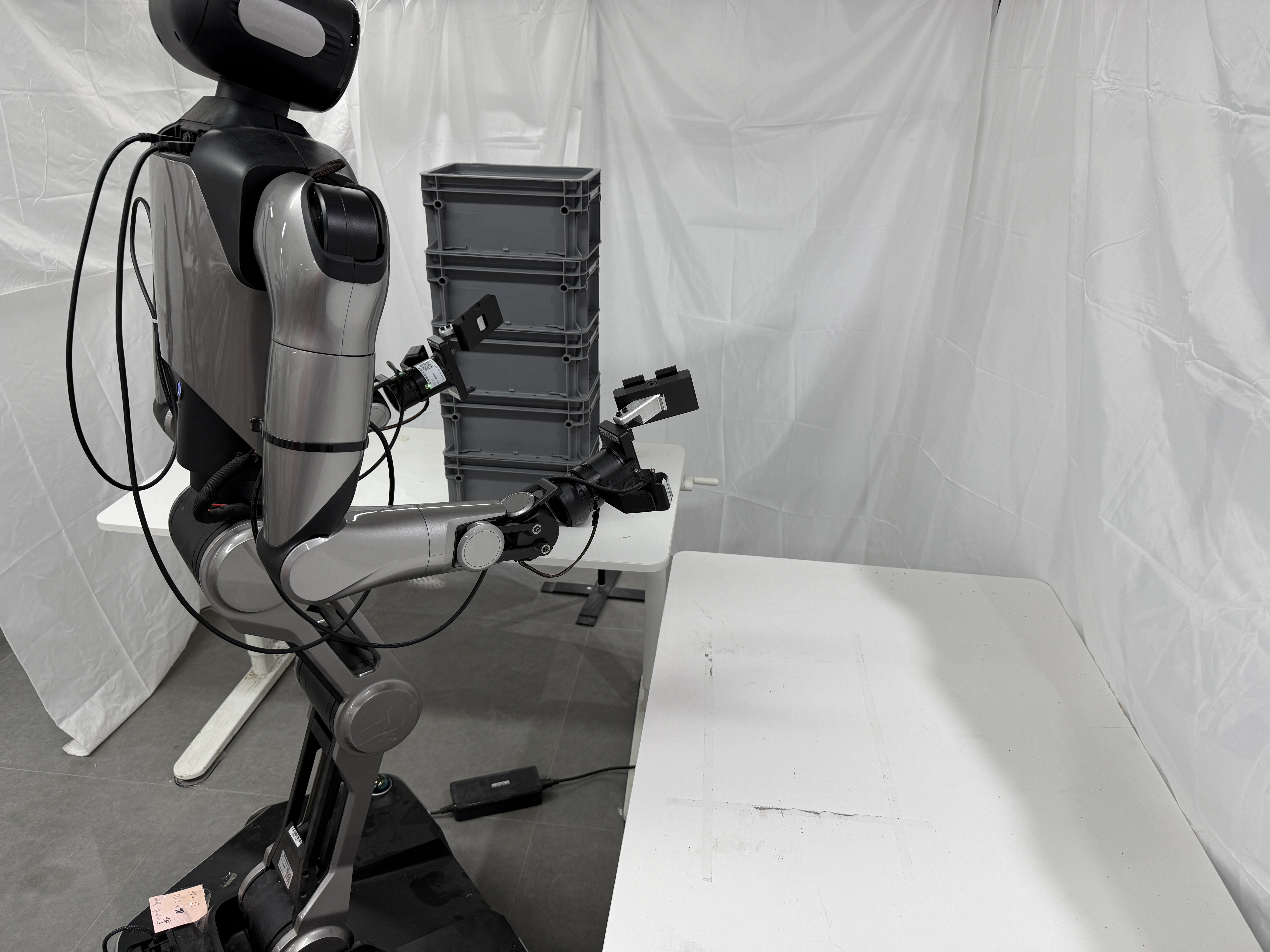}
        \caption{\RonePro: Box Transfer and Stacking}
    \end{subfigure}

    \caption{\textbf{Overview of real-world fine-tuning evaluation settings.} We evaluate four manipulation tasks instantiated as six task-embodiment settings across \RonePro and \RoneLite. The \RonePro settings include towel folding, carton folding, and box transfer and stacking, while the \RoneLite settings include towel folding, carton folding, and pencil-case packing. Towel folding and carton folding are shared across both embodiments, enabling cross-embodiment comparison under the same task objectives.}
    \label{fig:gbench_tasks}
    \vspace{-0.5em}
\end{figure*}

Fig.~\ref{fig:gbench_tasks} shows the six real-world evaluation settings across the two robot embodiments.
The four tasks cover deformable object manipulation, contact-rich assembly, sequential object interaction, and whole-body bimanual coordination.

\begin{itemize}[leftmargin=*]

\item \textbf{Folding Towel.}
This task is evaluated on both \RoneLite and \RonePro.
The robot is required to 1) grasp a towel from a basket, 2) lift and unfold the towel through coordinated bimanual motion, 3) flatten the towel on the tabletop, 4) fold the towel into a predefined configuration, and 5) place the folded towel into a designated target area.
This task is challenging because towels exhibit highly deformable and unstable geometric states during manipulation.
Small errors in grasping or tension control can accumulate throughout the folding process, leading to misalignment, incomplete folds, or entanglement.
Successful execution therefore requires accurate dual-arm coordination, continuous shape regulation, and long-horizon manipulation of deformable objects.

\item \textbf{Folding Carton.}
This task is evaluated on both \RoneLite and \RonePro.
The robot is required to transform a flat carton into a complete box structure through a sequence of predefined folding operations.
The task involves multiple stages of coordinated bimanual interaction, including edge alignment, surface folding, and structure stabilization.
Since the carton is non-rigid and sensitive to manipulation errors, minor inaccuracies during intermediate folding stages may damage the structure or prevent successful assembly.
The task therefore demands precise dual-arm coordination, accurate contact control, and stable sequential execution.

\item \textbf{Box Transfer and Stacking.}
This task is evaluated on \RonePro.
The robot is required to sequentially transfer five boxes from one table to another and stack them into a stable configuration.
During placement, each box must be accurately aligned with the grooves of the box below.
Unlike tabletop-only manipulation tasks, this task requires coordinated control of the upper body, including both arms and the torso, to achieve sufficient reachability and stable motion during transportation and placement.
The main challenge lies in precise spatial alignment during stacking, as small positioning errors can cause instability or collapse of the stack.

\item \textbf{Pencil-Case Packing.}
This task is evaluated on \RoneLite.
The robot is required to 1) unzip a pencil case from its closed state, 2) identify and sequentially place designated stationery items into the pencil case, and 3) close the zipper after all target objects have been inserted.
This task combines deformable object manipulation with fine-grained tool interaction.
The zipper is small and requires accurate manipulation to operate reliably, while the deformable structure of the pencil case introduces additional geometric uncertainty during interaction.
In addition, the robot must identify target objects and perform sequential pick-and-place operations under cluttered tabletop conditions.

\end{itemize}

For all tasks, we report both task success rate and process score.
The process score evaluates intermediate task progress based on predefined stage-wise completion criteria, enabling finer-grained comparison between different policies when a task is only partially completed.

\begin{figure}[thpb]
    \centering

    \includegraphics[width=0.95\columnwidth]{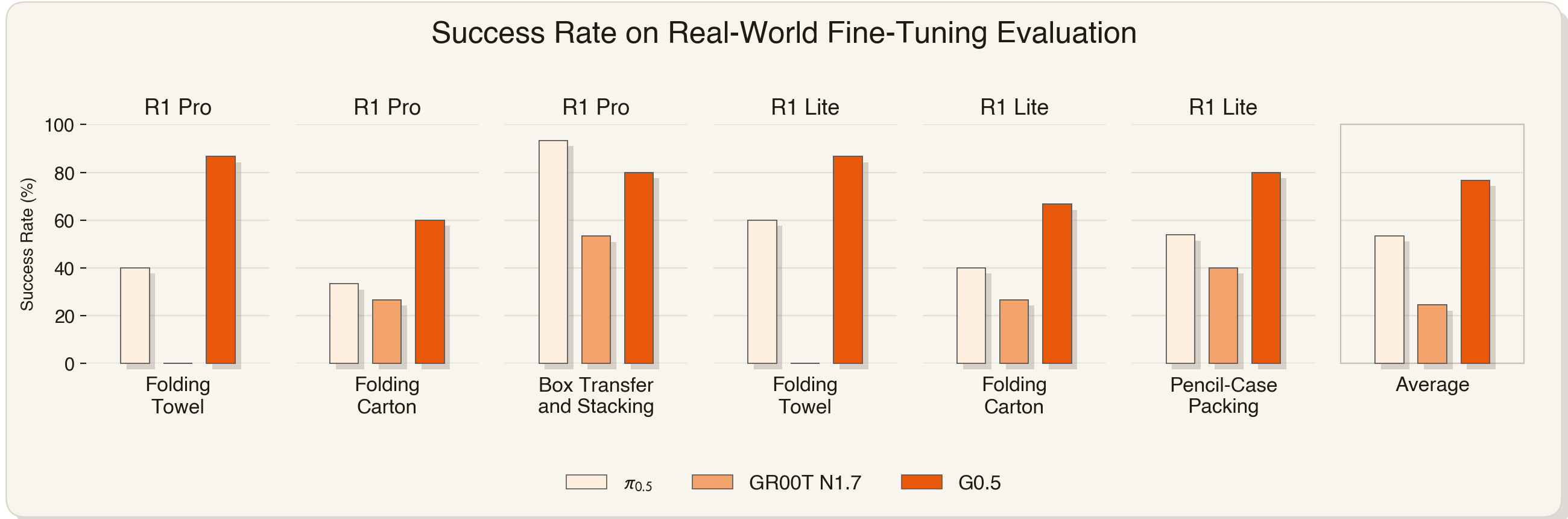}
    \vspace{0.5em}

    \includegraphics[width=0.95\columnwidth]{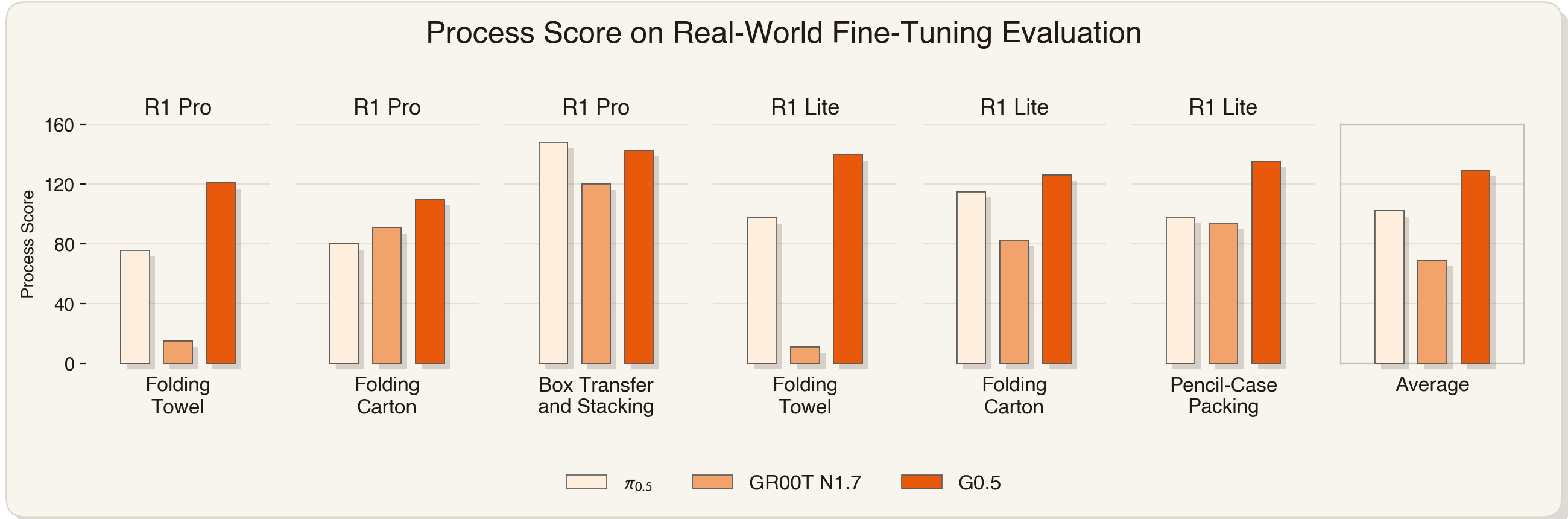}

    \caption{\textbf{Performance on real-world fine-tuning evaluation.} We evaluate \Gm against $\pi_{0.5}$ and GR00T-N1.7 on four manipulation tasks instantiated as six task-embodiment settings across \RonePro and \RoneLite. \Gm achieves strong overall performance in both task success rate and process score, demonstrating robust long-horizon manipulation capability across different embodiments.}
    \label{fig:gbench_results}
    \vspace{-0.5em}
\end{figure}

\subsubsection{Quantitative Results}

Fig.~\ref{fig:gbench_results} summarizes the quantitative results across the six real-world evaluation settings.
Overall, \Gm achieves the highest average performance among the three models.
Across all six settings, \Gm obtains an average success rate of 76.7\%, compared with 53.3\% for $\pi_{0.5}$ and 24.4\% for GR00T-N1.7.
\Gm also achieves an average process score of 129.2, compared with 105.2 for $\pi_{0.5}$ and 68.9 for GR00T-N1.7.

\Gm achieves the highest success rate in five out of the six settings.
The only exception is the \RonePro box transfer and stacking setting, where $\pi_{0.5}$ achieves a higher final success rate of 93.3\%, while \Gm achieves 80.0\%.
However, \Gm remains competitive in this setting, achieving a process score of 142.5 compared with 148.0 for $\pi_{0.5}$, and outperforming GR00T-N1.7 in both success rate and process score.

The two shared tasks, towel folding and carton folding, allow direct comparison across the two robot embodiments.
On these four shared task-embodiment settings, \Gm achieves an average success rate of 75.0\%, outperforming $\pi_{0.5}$ at 43.3\% and GR00T-N1.7 at 13.3\%.
For process score on the same shared settings, \Gm achieves an average score of 124.3, compared with 92.0 for $\pi_{0.5}$ and 49.9 for GR00T-N1.7.
These results indicate that \Gm maintains strong performance not only on embodiment-specific tasks, but also on the same task objectives executed by different robot configurations.

The performance of \Gm is also balanced across embodiments.
On \RonePro, \Gm achieves an average success rate of 75.6\% and an average process score of 124.5.
On \RoneLite, \Gm achieves an average success rate of 77.8\% and an average process score of 133.8.
This suggests that \Gm adapts effectively to both the 6-DoF dual-arm embodiment of \RoneLite and the 7-DoF humanoid upper-body embodiment of \RonePro under the same fine-tuning and evaluation protocol.

\subsection{Pick-and-Place Benchmark}
\label{subsec:pp_bench}
Large-scale pretraining is expected to improve not only low-level action generation, but also language following in visually cluttered scenes. In real-world manipulation, a policy must first identify the object and target specified by the instruction before executing primitive skills such as picking and placing. This distinction is critical because failures may arise either from incorrect language following, where the robot interacts with the wrong object, or from low-level execution errors after the correct target has been selected. To disentangle these factors, we introduce the Pick-and-Place Benchmark (PP Bench), which separately reports language following rate and final task success rate.
\paragraph{Dataset and Post-training Setup.}
We collect 50 hours of tabletop manipulation data using the \RoneLite robot. Each scene contains 5--20 objects randomly placed on the table with varying positions and orientations. The robot is instructed to pick up a specified object and place it into a specified container. Each instruction explicitly specifies both the target object and the target container, following the same format used during pretraining, e.g., \emph{``Task: Pick up the yellow utility knife with the left hand and place it into the white basket.''}

To evaluate the effect of post-training scale, we construct three nested subsets from the 50-hour dataset: 1H, 10H, and 50H, where the 1H subset is sampled from the 10H subset and the 10H subset is sampled from the full dataset. The models are trained for 12, 8, and 4 epochs on the 1H, 10H, and 50H subsets, respectively. This setup allows us to compare different data scales while keeping the data distribution consistent across subsets.

\begin{figure}[thpb]
\centering
\includegraphics[width=\columnwidth]{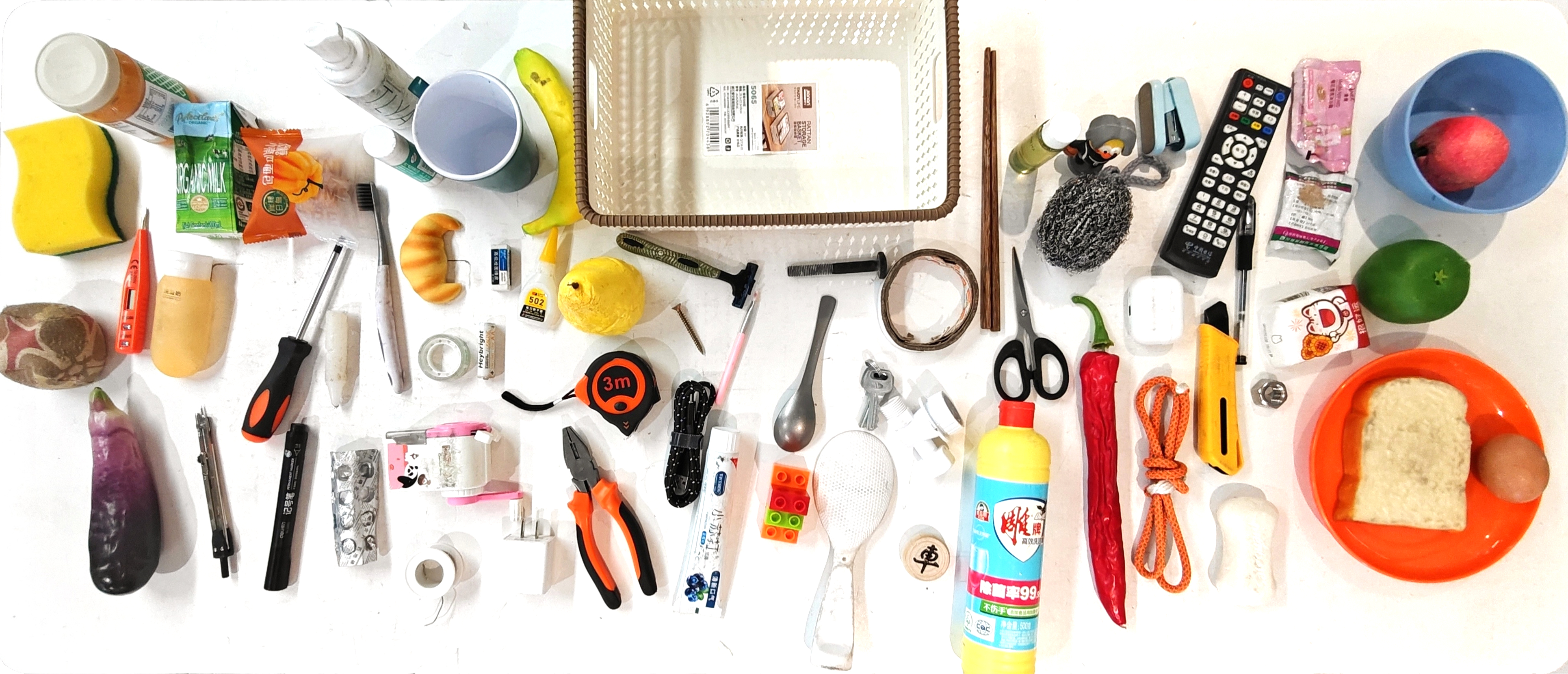}
\caption{\textbf{Pick-and-Place Benchmark setting.} The evaluation set contains 64 object categories and 3 container categories. Each trial presents 16 randomly arranged objects and containers, requiring the robot to identify the instructed target among distractors and place it into the specified container. Each policy is evaluated across all object categories in real-world trials to reduce variance.}
\label{fig:ppbench_setup}
\end{figure}

\paragraph{Evaluation Setting.}
The test set includes both in-distribution and out-of-distribution object categories. We randomly sample 48 objects from the 50H subset and additionally include 16 categories absent from all post-training data for out-of-distribution evaluation. The full set of benchmark objects and containers is shown in Fig.~\ref{fig:ppbench_setup}. During evaluation, each tabletop scene contains 16 randomly placed objects and containers, following the same setup as data collection. In each trial, the robot receives a language instruction and is required to place the specified object into the specified container. Each model is evaluated over 64 real-world trials to reduce variance.
For fair comparison, different models are evaluated on the same instruction under identical object layouts, container placements, and robot initial states. This paired evaluation protocol reduces variance from scene configuration and isolates the effect of the policy.

We report two metrics: language following rate and task success rate. Language following rate measures whether the robot selects the object specified by the instruction among distractors. A trial is counted as language-following success if the robot moves toward the target object and attempts to grasp it. Task success rate measures whether the robot completes the full instruction by successfully grasping the specified object and placing it into the specified container.

\begin{figure}[thpb]
    \centering
    \includegraphics[width=\columnwidth]{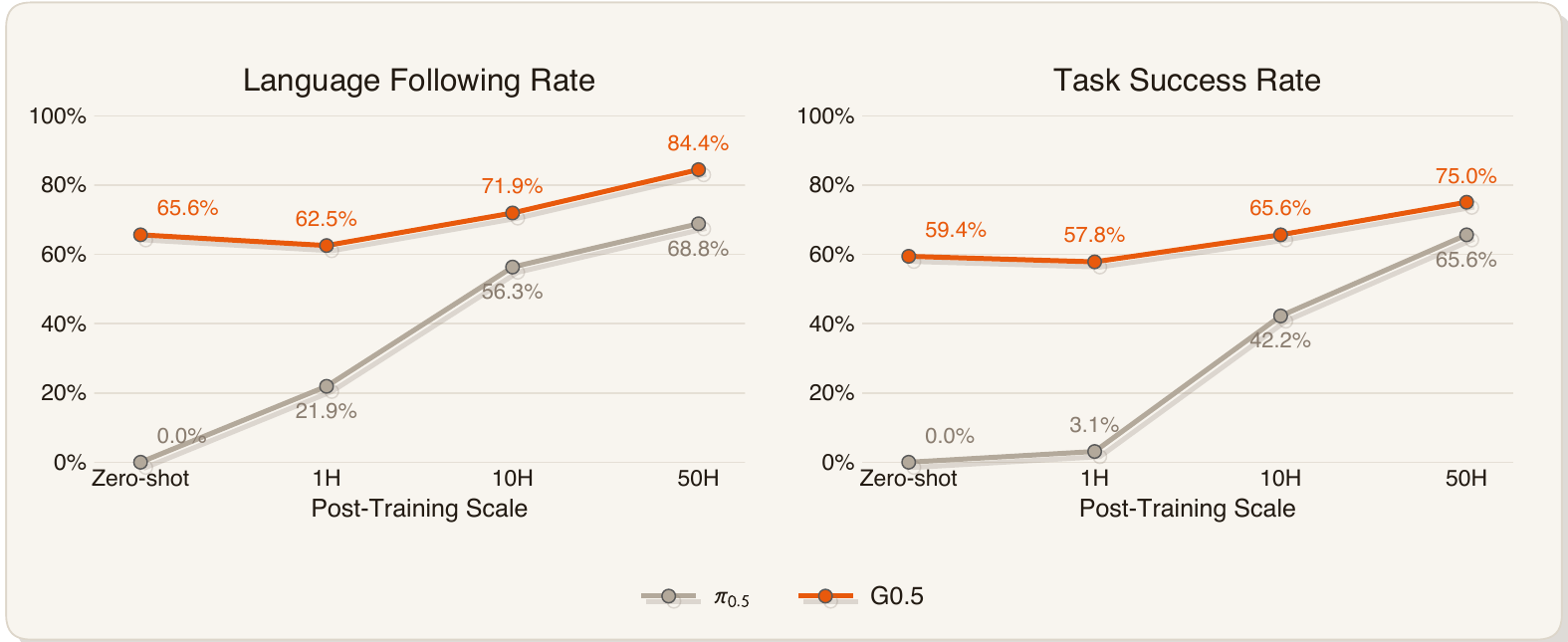}
    \caption{\textbf{PP Bench evaluation results.} Left: language following rate. Right: task success rate. We compare \Gm and $\pi_{0.5}$ across zero-shot, 1H, 10H, and 50H post-training settings.}
    \label{fig:ppbench_results}
\end{figure}

\paragraph{Results and Analysis.}
Using PP Bench, we evaluate \Gm from four perspectives: zero-shot capability, the effect of post-training, comparison with $\pi_{0.5}$, and the benefit of additional context (Fig.~\ref{fig:ppbench_results}).

\vspace{0.3em}
\noindent\textit{Zero-shot capability.}
\Gm exhibits strong zero-shot language following. Without any PP-specific post-training data, it achieves a language following rate of 65.6\% and a task success rate of 59.4\%. This indicates that large-scale pretraining provides transferable instruction-following ability and basic pick-and-place action priors.

\vspace{0.3em}
\noindent\textit{Effect of post-training.}
Post-training further improves both semantic grounding and execution reliability. \Gm reaches language following rates of 62.5\%, 71.9\%, and 84.4\% under the 1H, 10H, and 50H settings, respectively, with corresponding task success rates of 57.8\%, 65.6\%, and 75.0\%. The consistent improvement from 10H to 50H suggests that additional post-training data strengthens grounding for in-distribution objects and improves action reliability on the \RoneLite embodiment.

\vspace{0.3em}
\noindent\textit{Comparison with $\pi_{0.5}$.}
Compared with $\pi_{0.5}$, \Gm achieves higher language following and task success across all post-training scales. The gap is most pronounced in the zero-shot and 1H settings, where $\pi_{0.5}$ shows limited transfer to the \RoneLite setup. With 50H post-training, $\pi_{0.5}$ improves to 68.8\% in language following and 65.6\% in task success, indicating that target-domain post-training helps adapt the model to the instruction format, object-container distribution, and \RoneLite action interface. Nevertheless, under the same 50H setting, \Gm still outperforms $\pi_{0.5}$ by 15.6 percentage points in language following and 9.4 percentage points in task success. We attribute this advantage to large-scale web-data co-training, which improves open-vocabulary semantic understanding, and to robot pretraining that includes the \RoneLite embodiment, which provides stronger action priors and execution quality.

\vspace{0.3em}
\noindent\textit{Additional referring context.}
\begin{table}[t]
\centering
\small
\caption{\textbf{Ablation on referring context for PP Bench.}
We evaluate \Gm with progressively richer \emph{referring context}, i.e., auxiliary inputs appended to the standard instruction that help the policy identify the language-specified target object beyond its category name.}
\label{tab:ppbench_context_ablation}
\begin{tabular}{lccc}
\toprule
Metric
& Name-only
& + Box Coord.
& + Target Visual \\
\midrule
Language Following Rate
& 84.4\%
& 85.9\%
& 98.4\% \\
Task Success Rate
& 75.0\%
& 76.6\%
& 84.4\%  \\
\bottomrule
\end{tabular}
\end{table}
Failure analysis shows that many remaining language-following errors occur on long-tail objects that are visually ambiguous, partially occluded, or difficult to identify from language alone. We therefore evaluate several referring-context variants in Table~\ref{tab:ppbench_context_ablation}.
\emph{Name-only} uses the standard instruction format, where the prompt specifies only the target object and container names.
\emph{+ Box Coord.} augments the instruction with textual coordinate tokens for the target object and container boxes. Despite providing explicit spatial cues, this setting does not improve over the name-only baseline. We hypothesize that directly injecting coordinate tokens introduces an instruction-format shift from the name-only robot pretraining distribution, and that the post-training data may be insufficient for the action head to reliably exploit this new conditioning signal.
\emph{+ Target Visual} further augments the input with cropped visual regions of the target object and container. These visual contexts are appended to the original camera views and encoded with the same visual encoder as the standard observations. Compared with box coordinates alone, the target visual context provides fine-grained local appearance cues, such as texture, shape, and category-specific visual details. These cues are particularly useful for targets that are difficult to disambiguate from language alone or belong to long-tail categories with limited linguistic exposure but distinctive visual appearance, such as a Chinese chess piece labeled ``horse''. This setting significantly improves language following to 98.4\% and task success to 84.4\%.

\subsection{Zero-Shot Probe of CoT and Action Head}
\label{subsec:cot_head}

\begin{figure}[t]
\centering
\includegraphics[width=\linewidth]{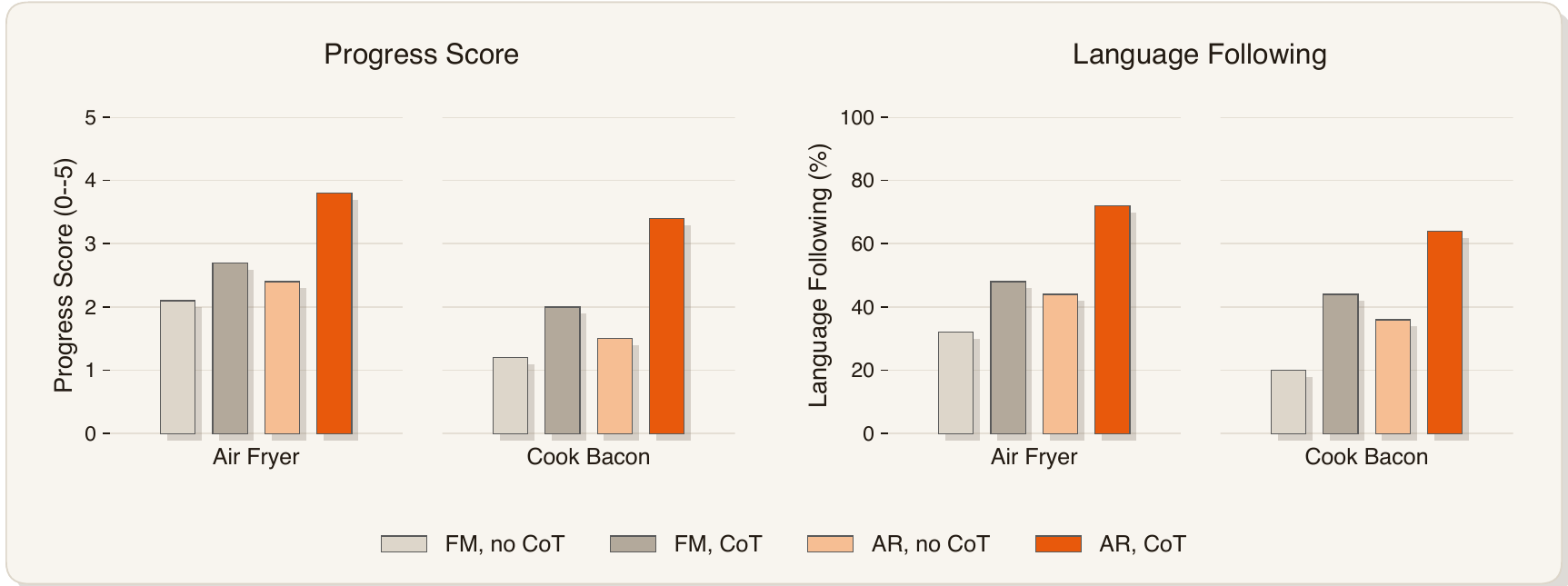}
\caption{Long-horizon zero-shot probe on \emph{Air Fryer} and \emph{Cook Bacon} (same pretrained \Gm checkpoint with only inference-time switching). Progress score (0--5, number of completed sub-stages) and language-following rate across the four decoder $\times$ CoT cells: both metrics show the same ordering, with AR$+$CoT clearly leading on each task.}
\label{fig:cot_head_long}
\end{figure}

We isolate the contributions of chain-of-thought and the action decoder by taking a single pretrained \Gm checkpoint and varying \emph{only the inference-time configuration}; no parameters are fine-tuned or adapted, and the same weights are reused across all cells. The probe spans three tasks of increasing horizon on the \RoneLite platform: PP~Bench (single stage), and two new long-horizon zero-shot tasks---\emph{Air Fryer} and \emph{Cook Bacon}---each decomposed into five sequential stages (\emph{Air Fryer:} approach $\to$ open door $\to$ grasp bread $\to$ place inside $\to$ close door; \emph{Bacon:} approach $\to$ grasp bacon $\to$ place in pan $\to$ turn on stove $\to$ flip). At runtime the policy receives a per-stage natural-language instruction for the current sub-goal. We toggle (i)~the decoder---\emph{AR} for autoregressive action tokens versus \emph{FM} for an additional flow-matching head (optional at inference)---and (ii)~the chain-of-thought stream, where the model optionally emits subtask and bounding-box reasoning before each action. When CoT is on, both decoders read from the post-CoT hidden state, so the comparison isolates the decoding interface rather than the conditioning input. We report the language-following rate and, for the long-horizon tasks, a progress score equal to the number of completed sub-stages (0--5). PP~Bench follows the zero-shot evaluation protocol of Sec.~\ref{subsec:pp_bench} ($64$ rollouts); the long-horizon tasks (Fig.~\ref{fig:cot_head_long}) use $n{=}5$ rollouts per cell.

\paragraph{Finding 1: CoT improves grounding and execution within long-horizon stage-conditioned rollouts.}
On the single-stage PP~Bench, CoT brings essentially no change to language following: AR moves from $65.6$ to $67.2$ and FM from $59.4$ to $60.9$ over 64 rollouts---at most $\sim$1.6 percentage points for either decoder. With only one grounding event per rollout, there is little room for per-stage reasoning to help. On the five-stage Air~Fryer and Bacon tasks the picture is different: with per-stage sub-goals supplied at runtime, CoT lets the policy ground the relevant object and execute each sub-step more reliably. With this structure the AR head's progress score lifts from $2.4$ to $3.8$ on Air~Fryer and from $1.5$ to $3.4$ on Bacon, with the language-following rate rising in step (Fig.~\ref{fig:cot_head_long}). The benefit thus appears where the task is presented as a sequence of stage-conditioned sub-goals. Air~Fryer and Bacon are household manipulation scenes that \emph{do not appear in the pretraining data}; CoT improves per-stage grounding and execution on them without any retraining.

\paragraph{Qualitative observation: instruction wording.}
Beyond toggling CoT, we informally observed that the exact wording of the per-stage instruction affects behavior on these tasks. Two patterns recurred. First, adding adverbial or spatial qualifiers (e.g., \emph{``push it in hard''}, \emph{``vertically''}) tended to bring the executed motion closer to the intended sub-goal. Second, a single physical action often admits several near-synonymous verbs---closing the air-fryer drawer can be phrased as \emph{press}, \emph{push in}, or \emph{close}---and substituting among them changed which rollouts succeeded in our small set. We report these only as qualitative observations. This probe is AR-only.

\paragraph{Finding 2: AR appears to follow the CoT more closely than the FM head.}
Under matched CoT, the AR head benefits more than the FM head (Fig.~\ref{fig:cot_head_long}): on Air~Fryer, CoT lifts AR's progress score from $2.4$ to $3.8$ while FM moves only from $2.1$ to $2.7$; on Bacon, $1.5\!\to\!3.4$ for AR versus $1.2\!\to\!2.0$ for FM. The language-following rate echoes this gap under matched CoT (Air~Fryer 72 vs.\ 48; Bacon 64 vs.\ 44). We \emph{hypothesize} that this reflects the decoding interface rather than the reasoning content: the autoregressive action tokens are emitted in the same stream as the CoT and can attend to it directly, whereas the FM head conditions on a pooled summary of the hidden state. We do not directly probe this mechanism and leave its verification to future work.

\paragraph{CoT correctness.}
The CoT is generated autoregressively by the VLA; once it terminates, the action is either continued autoregressively (AR) or sampled by the FM head. AR and FM are therefore run as two separate rollouts whose CoT traces, while produced by the same mechanism, are not identical, since the rollouts diverge once actions are executed. We hand-scored subtask text and bounding-box correctness on the CoT-on rollouts for both heads and found comparable quality---roughly $90\%$ on PP~Bench, $85\%$ on Air~Fryer, and $80\%$ on Bacon, with no systematic AR--FM difference. The progress-score gap between the two heads under matched CoT therefore does not appear to come from differences in reasoning quality, which is consistent with the decoding-interface hypothesis in Finding~2.

\subsection{\Gm Fine-Tunes with RL Out of the Box}
\label{subsec:rl_grpo}

An autoregressive policy emits actions as tokens, so it exposes exact token-level log-probabilities---precisely the quantity that ratio-based RL algorithms consume. A flow-matching head offers no such likelihood and must be reformulated before the same algorithms apply. We therefore ask whether the autoregressive interface is the easier one to optimize with RL, and test this in a low-data regime where post-training gains come almost entirely from exploration rather than imitation. We first jointly post-trained the AR and FM policies using a single demonstration trajectory per task on LIBERO \citep{libero}. We then selected four tasks on which the two policies reached comparable initial success rates and applied Group Relative Policy Optimization (GRPO) \citep{shao2024deepseekmath}. For the AR policy, actions were sampled with a temperature to encourage stochastic exploration while retaining tractable token-level log-probabilities. For the FM policy, we introduced an SDE into the denoising procedure and treated the denoising trajectory as a Markov process in order to approximate the policy log-probability, following RLinf \citep{yu2025rlinf}.

\begin{figure}[t]
\centering
\includegraphics[width=0.82\linewidth]{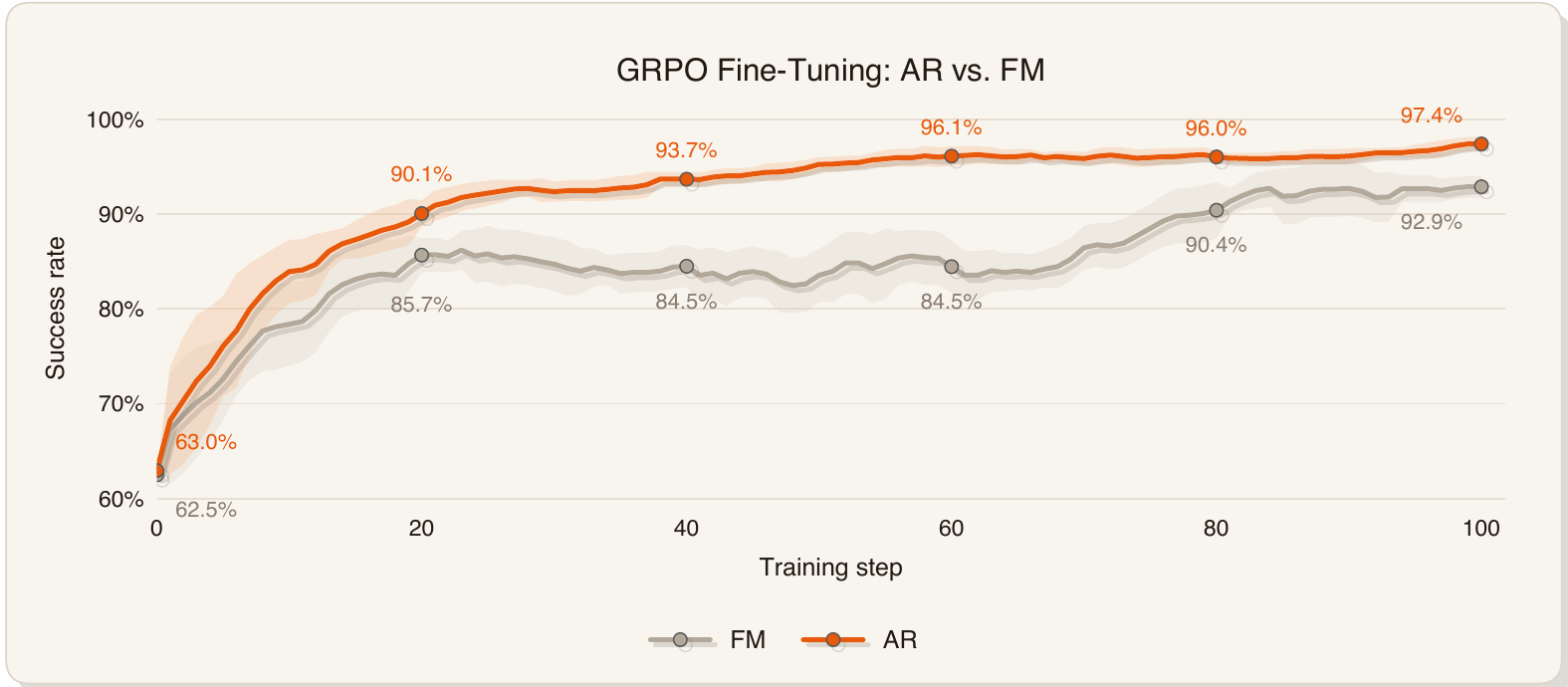}
\caption{\textbf{GRPO fine-tuning: AR vs.\ FM.} Average success rate over four LIBERO tasks during GRPO fine-tuning from a single demonstration per task (mean $\pm$ std over seeds). Starting from AR and FM policies with comparable initial success rates, the AR policy converges faster, attains a higher final success rate, and shows lower run-to-run variance.}
\label{fig:rl_grpo}
\end{figure}

As shown in Fig.~\ref{fig:rl_grpo}, the AR policy converges substantially faster, reaches a higher final success rate, and trains more stably, with lower variance, than the FM policy. We hypothesize that this advantage stems from the AR policy's native and direct likelihood parameterization, which makes policy probability ratios straightforward to compute. Applying ratio-based policy optimization to the FM policy, in contrast, requires an auxiliary stochastic reformulation, a discretization of the denoising dynamics, and additional noise-schedule choices; these components may introduce greater optimization sensitivity and gradient variance.

\section{Conclusion}

We have argued, and empirically supported, that the path forward for VLA models is to let the VLM be what it was pretrained to be---an autoregressive reasoner that now also acts, remembers, and adapts in-context---rather than to design ever more sophisticated action experts on top of an underutilized backbone. \Gm instantiates this commitment with a single cross-entropy objective over a shared vocabulary, supported by a cross-embodiment action codec, a native chain-of-thought stream, and a multi-second visual memory module.

Three observations point to a structural rather than incidental advantage of the autoregressive route. First, on Pick-and-Place benchmark, the zero-shot language-following rate of \Gm exceeds the post-trained $\pi_{0.5}$ baseline, suggesting that AR action supervision protects---rather than degrades---the VLM's instruction-following ability. Second, on the 2025 BEHAVIOR Challenge, a single \Gm checkpoint trained for only one post-training epoch surpasses both $\pi_{0.5}$ post-trained for four epochs and the four-checkpoint winner, indicating that the pretrained representations carry generalist mobile manipulation priors. Third, when fine-tuned under matched compute and an identical protocol on the \RoneLite and \RonePro platforms, \Gm beats $\pi_{0.5}$ and GR00T-N1.7, indicating that the advantage over VLM-as-encoder architectures persists under apples-to-apples conditions rather than reflecting differences in training budget.

\Gm inherits two acknowledged failure modes that future work should address. Drawer-insertion and semi-transparent cabinet tasks remain weak across both \Gm variants, pointing to a sensing limit not closed by AR alone; and our visual memory captures only seconds of history, leaving long-horizon memory open to research. Lower-body actuation is represented in the unified action space, but is not evaluated separately in this work. More broadly, the prompt-level controllability our zero-shot probe begins to surface---where per-stage instruction wording shifts AR$+$CoT rollouts on out-of-distribution household tasks---deserves a systematic empirical study of its own. We hope the released pretrained backbone serves as a starting point for further work in these directions.

\section{Contributors}



\smallskip
\noindent\textbf{Data engineering:}
Tao Jiang, Ke Dong, Xiaoshu Ren, Chenru Wu, Xiao Liu, Tianyuan Yuan, Zibin Dong, Zihan Guo, Zijie Zhao

\smallskip
\noindent\textbf{Annotation \& supplemental data:}
Tianyuan Yuan, Tao Jiang, Changxun Pan, Xinlei Zhang, Chenru Wu, Haonan Liu, Haodong Yang, Bowen Zhang

\smallskip
\noindent\textbf{Policy training \& research:}
Yicheng Liu, Zibin Dong, Tianyuan Yuan, Baijun Ye, Shicheng Cao, Shaoting Zhu, Xiao Liu

\smallskip
\noindent\textbf{Policy evaluation \& benchmarking:}
 Anqi Yang, Zihan Guo, Baijun Ye, Zibin Dong, Tao Jiang, Yue Sun, Tianyuan Yuan, Tailai Cheng, Changxun Pan, Jianning Cui, Shicheng Cao, Haonan Liu, Shaoting Zhu, Zijie Zhao, Haoyu Zhang, Jiahui Niu, Shiduo Zhang

\smallskip
\noindent\textbf{Writing \& illustration:}
Yicheng Liu, Zibin Dong, Baijun Ye, Shicheng Cao, Haonan Liu, Tailai Cheng, Tao Jiang, Zihan Guo, Anqi Yang, Yue Sun, Tianyuan Yuan, Shiduo Zhang, Hang Zhao

\smallskip
\noindent\textbf{Project lead:} Yicheng Liu \quad\textbullet\quad \textbf{Project PI:} Hang Zhao

\newpage
\bibliographystyle{unsrtnat}
\bibliography{galaxea}


\clearpage
\appendix

\section{Appendix / supplemental material}

\begin{table*}[h!]
\centering
\small
\caption{Detailed results on the 2025 BEHAVIOR Challenge (50 tasks, 10 instances each). \textit{Task Success Score} is the challenge ranking metric (task progress). The first place solution by the Robot Learning Collective (RLC)~\citep{1st_place} uses a set of 4 checkpoints; $\pi_{0.5}$~\citep{pi0.5} (4 epochs) and \Gm each use a single checkpoint, averaged over two eval runs. Best/second best in \textbf{bold}/\underline{underline}.}
\label{tab:b1k_results}
\resizebox{0.92\textwidth}{!}{
\begin{tabular}{l ccccc}
\toprule
Task & RLC~\citep{1st_place} & Comet~\citep{comet} & $\pi_{0.5}$ (4 epochs)~\citep{pi0.5} & \Gm (1 epoch) & \Gm (4 epochs) \\
\midrule
assembling gift baskets & 0.2125 & 0.0000 & 0.2312 & \best{0.5188} & \underline{0.4938} \\
attach camera to tripod & 0.0000 & 0.0000 & 0.0000 & 0.0000 & 0.0000 \\
boxing books for storage & 0.0000 & 0.0000 & 0.0000 & 0.0000 & 0.0000 \\
bringing in wood & 0.0667 & \best{0.5000} & 0.1667 & 0.3333 & \underline{0.3500} \\
bringing water & 0.2667 & \best{0.9000} & 0.6500 & \underline{0.8333} & 0.7000 \\
can meat & 0.0000 & 0.0000 & \underline{0.0333} & 0.0111 & \best{0.0444} \\
canning food & 0.0100 & 0.0000 & 0.0550 & \underline{0.0800} & \best{0.1100} \\
carrying in groceries & \underline{0.1500} & 0.0000 & \best{0.1750} & 0.0750 & \underline{0.1500} \\
chop an onion & 0.3000 & 0.0000 & 0.1750 & \best{0.4125} & \underline{0.4000} \\
chopping wood & 0.1000 & 0.0000 & 0.0875 & \best{0.2375} & \underline{0.2000} \\
clean a patio & 0.0000 & 0.0000 & 0.0000 & 0.0000 & 0.0000 \\
clean a trumpet & 0.0000 & 0.0000 & 0.0000 & 0.0000 & 0.0000 \\
clean boxing gloves & \underline{0.2000} & 0.0000 & 0.0000 & \best{0.2250} & 0.1750 \\
clean desk & 0.1273 & 0.0000 & 0.2227 & \best{0.2864} & \underline{0.2591} \\
clean plates and food & 0.1857 & 0.0000 & \best{0.3857} & 0.1786 & \underline{0.1929} \\
clear food to fridge & 0.0800 & 0.0000 & 0.0800 & \underline{0.1800} & \best{0.2700} \\
collect childrens toys & 0.4333 & 0.0000 & 0.4214 & \underline{0.5857} & \best{0.5929} \\
cook bacon & \best{0.7571} & 0.0000 & \underline{0.4214} & 0.0714 & 0.3214 \\
cook cabbage & 0.0000 & 0.0000 & 0.0500 & \underline{0.1500} & \best{0.2000} \\
cook hot dogs & 0.8500 & \best{1.0000} & \underline{0.9250} & 0.4500 & 0.9000 \\
freeze pies & 0.0143 & \best{0.1571} & \underline{0.1357} & 0.0571 & 0.0429 \\
get organized for work & 0.0200 & 0.0000 & 0.0200 & \underline{0.1050} & \best{0.1100} \\
hanging pictures & 0.0000 & \best{0.2000} & 0.0000 & 0.0000 & 0.0000 \\
hiding Easter eggs & \underline{0.1444} & \best{0.2444} & 0.0778 & 0.0500 & 0.0500 \\
loading the car & 0.2000 & 0.0000 & 0.0000 & \underline{0.2333} & \best{0.2833} \\
make microwave popcorn & \underline{0.9000} & 0.7000 & \best{0.9500} & 0.1500 & 0.5500 \\
make pizza & 0.0000 & 0.0000 & 0.0000 & 0.0000 & 0.0000 \\
move boxes to storage & \underline{0.6500} & \best{1.0000} & 0.2000 & 0.6250 & 0.5500 \\
outfit basic toolbox & \underline{0.2571} & 0.1000 & \best{0.3429} & 0.1429 & 0.2500 \\
pick up toys & 0.3000 & 0.0000 & 0.1833 & \underline{0.3167} & \best{0.4083} \\
pick up trash & 0.6667 & 0.7667 & 0.5500 & \underline{0.8167} & \best{0.8500} \\
prepare lunch box & 0.5167 & 0.0000 & 0.5417 & \best{0.5667} & \best{0.5667} \\
put away Halloween decor. & 0.2000 & \underline{0.5000} & 0.3714 & 0.4857 & \best{0.5286} \\
put dishes away & \underline{0.2714} & 0.0000 & \best{0.5250} & 0.1393 & 0.2464 \\
put shoes on rack & 0.5000 & 0.5400 & 0.3650 & \best{0.6450} & \underline{0.5650} \\
put up Christmas decor. & 0.4333 & 0.0000 & \best{0.5611} & \underline{0.4667} & 0.3889 \\
rearrange kitchen furn. & 0.3000 & \underline{0.3750} & \best{0.3875} & 0.3500 & 0.3625 \\
set up coffee station & 0.1500 & 0.2167 & \best{0.2833} & \underline{0.2500} & 0.1917 \\
setting mousetraps & 0.3333 & 0.0000 & 0.1083 & \underline{0.5083} & \best{0.5667} \\
setting the fire & \best{0.3250} & 0.2000 & 0.0750 & \underline{0.3125} & 0.1250 \\
slicing vegetables & 0.1889 & 0.0000 & 0.2278 & \best{0.4444} & \underline{0.2611} \\
sorting household items & 0.0625 & 0.0000 & \underline{0.1938} & 0.1437 & \best{0.2062} \\
sorting vegetables & 0.4769 & 0.0000 & \underline{0.6000} & 0.5231 & \best{0.6231} \\
spraying for bugs & \best{0.2500} & 0.1000 & 0.1000 & \underline{0.2000} & 0.1500 \\
spraying fruit trees & \underline{0.3000} & \best{0.3500} & 0.2000 & 0.1250 & 0.1500 \\
storing food & 0.3750 & 0.0000 & \underline{0.5750} & 0.4938 & \best{0.6625} \\
tidying bedroom & 0.4000 & 0.0000 & 0.3500 & \underline{0.5667} & \best{0.6167} \\
turning on radio & \underline{0.6000} & \best{1.0000} & 0.1500 & 0.3000 & 0.0500 \\
wash a baseball cap & 0.4500 & 0.3000 & 0.6000 & \underline{0.6500} & \best{0.7500} \\
wash dog toys & 0.0000 & 0.0000 & \best{0.3750} & \underline{0.2250} & 0.2167 \\
\midrule
\textbf{Overall} & 0.2605 & 0.1830 & 0.2626 & \underline{0.2904} & \best{0.3136} \\
\bottomrule
\end{tabular}
}
\end{table*}

\begin{table*}[t]
\centering
\tiny
\caption{Per-task success rates of \Gm on RoboTwin 2.0 under clean and
randomized evaluation settings.}
\label{tab:robotwin_detail}
\setlength{\tabcolsep}{4pt}
\renewcommand{\arraystretch}{0.88}
\resizebox{0.46\textwidth}{!}{
\begin{tabular}{lcc}
\toprule
Task & Clean & Rand. \\
\midrule
Adjust Bottle & 100 & 100 \\
Beat Block Hammer & 100 & 96 \\
Blocks Ranking RGB & 100 & 100 \\
Blocks Ranking Size & 96 & 96 \\
Click Alarmclock & 100 & 100 \\
Click Bell & 100 & 100 \\
Dump Bin Bigbin & 94 & 96 \\
Grab Roller & 100 & 100 \\
Handover Block & 98 & 84 \\
Handover Mic & 98 & 100 \\
Hanging Mug & 40 & 56 \\
Lift Pot & 100 & 100 \\
Move Can Pot & 98 & 94 \\
Move Pillbottle Pad & 96 & 96 \\
Move Playingcard Away & 100 & 98 \\
Move Stapler Pad & 80 & 70 \\
Open Laptop & 100 & 98 \\
Open Microwave & 92 & 88 \\
Pick Diverse Bottles & 84 & 86 \\
Pick Dual Bottles & 96 & 88 \\
Place A2B Left & 94 & 94 \\
Place A2B Right & 94 & 90 \\
Place Bread Basket & 98 & 98 \\
Place Bread Skillet & 96 & 90 \\
Place Burger Fries & 100 & 98 \\
Place Can Basket & 68 & 72 \\
Place Cans Plasticbox & 96 & 100 \\
Place Container Plate & 100 & 98 \\
Place Dual Shoes & 88 & 94 \\
Place Empty Cup & 100 & 100 \\
Place Fan & 100 & 94 \\
Place Mouse Pad & 84 & 84 \\
Place Object Basket & 90 & 92 \\
Place Object Scale & 96 & 94 \\
Place Object Stand & 100 & 96 \\
Place Phone Stand & 94 & 98 \\
Place Shoe & 98 & 94 \\
Press Stapler & 92 & 92 \\
Put Bottles Dustbin & 90 & 90 \\
Put Object Cabinet & 90 & 90 \\
Rotate QRcode & 96 & 98 \\
Scan Object & 100 & 92 \\
Shake Bottle & 100 & 100 \\
Shake Bottle Horizontally & 100 & 100 \\
Stack Blocks Three & 98 & 100 \\
Stack Blocks Two & 100 & 100 \\
Stack Bowls Three & 94 & 86 \\
Stack Bowls Two & 96 & 90 \\
Stamp Seal & 88 & 92 \\
Turn Switch & 74 & 80 \\
\midrule
\textbf{Average} & \best{93.72} & \best{92.84} \\
\bottomrule
\end{tabular}
}
\end{table*}

\end{document}